\documentclass{article} 
\usepackage{iclr2026_conference,times}

\usepackage{amsmath,amsfonts,bm}

\def\eqref#1{equation~\ref{#1}}

\def\1{\bm{1}}

\DeclareMathAlphabet{\mathsfit}{\encodingdefault}{\sfdefault}{m}{sl}
\SetMathAlphabet{\mathsfit}{bold}{\encodingdefault}{\sfdefault}{bx}{n}

\usepackage{hyperref}
\usepackage{url}

\usepackage{graphicx}
\usepackage{amsmath}
\usepackage{amsthm}
\usepackage{amssymb}
\usepackage{algorithm}
\usepackage{algorithmic}
\usepackage{booktabs}
\usepackage{xcolor}
\iclrfinalcopy
\theoremstyle{definition}
\newtheorem{defn}{\protect\definitionname}
\theoremstyle{plain}
\newtheorem{theorem}{\protect\theoremname}
\theoremstyle{plain}
\newtheorem{lemma}{\protect\lemmaname}
\theoremstyle{plain}
\newtheorem{cor}{\protect\corollaryname}
\theoremstyle{plain}
\newtheorem*{lem*}{\protect\lemmaname}
\theoremstyle{plain}
\newtheorem*{thm*}{\protect\theoremname}
\theoremstyle{plain}
\newtheorem{fact}{\protect\factname}
\providecommand{\corollaryname}{Corollary}
\providecommand{\definitionname}{Definition}
\providecommand{\lemmaname}{Lemma}
\providecommand{\theoremname}{Theorem}
\providecommand{\factname}{Fact}

\title{An efficient, provably optimal algorithm \\for the 0-1 loss linear classification problem}

\author{Xi He \\
School of Computer Science\\
Peking University\\
Beijing, China \\
\texttt{xihe@pku.edu.cn} \\
\And
Max A. Little \\
School of Computational Neuroscience \\
University of Birmingham \\
Birmingham, UK \\
\texttt{maxl@mit.edu} \\
}

\begin{document}

\maketitle

\begin{abstract}
	Algorithms for solving the linear classification problem have a long
history, dating back at least to 1936 with linear discriminant analysis.
For linearly separable data, many algorithms can obtain the exact
solution to the corresponding 0-1 loss classification problem efficiently,
but for data which is not linearly separable, it has been shown that
this problem, in full generality, is NP-hard. Alternative approaches
all involve approximations of some kind, such as the use of surrogates
for the 0-1 loss (for example, the hinge or logistic loss), none of
which can be guaranteed to solve the problem exactly. Finding an efficient,
rigorously proven algorithm for obtaining an exact (i.e., globally
optimal) solution to the 0-1 loss linear classification problem remains
an open problem. 

By analyzing the combinatorial and incidence relations between hyperplanes and data points, we derive a rigorous construction algorithm, incremental cell enumeration (ICE),
that can solve the 0-1 loss classification problem exactly in $O\left(N^{D+1}\right)$---exponential
in the data dimension $D$. To the best of our knowledge, this is
the first standalone algorithm---one that does not rely on general-purpose
solvers---with rigorously proven guarantees for this problem. Moreover,
we further generalize ICE to address the polynomial hypersurface classification
problem in $O\left(N^{G+1}\right)$ time, where $G$ is determined by both the data dimension $D$ and the polynomial degree
$K$ defining the hypersurface. The correctness of our algorithm is
proved by the use of tools from the theory of hyperplane arrangements and
oriented matroids.

We demonstrate the effectiveness of our algorithm on real-world  datasets, achieving optimal training accuracy for small-scale  datasets and higher test accuracy on most  datasets. Furthermore, our complexity analysis shows that the ICE algorithm offers superior computational efficiency compared with state-of-the-art branch-and-bound algorithm.
\end{abstract}

\section{Introduction}

Increasingly, machine learning (ML) is being used
for high-stakes prediction applications that deeply impact human lives.
Many of these ML models are ``black boxes'' with highly complex, inscrutable
functional forms. In high-stakes applications such as healthcare and
criminal justice, black box ML predictions have incorrectly denied
parole \citep{wexler2017computer}, misclassified highly polluted
air as safe to breathe \citep{mcgough2018bad}, and suggested poor
allocation of valuable, limited resources in medicine and energy reliability
\citep{varshney2017safety}. In such high-stakes applications of ML,
we always want the best possible prediction, and we want to know how
the model makes these predictions so that we can be confident the
predictions are meaningful \citep{rudin-2022}. In short, the ideal
model is simple enough to be easily understood (\emph{interpretable}),
and optimally accurate (\emph{exact}).

Another compelling reason why simple models are preferable is because
such \emph{low complexity }models usually provide better \emph{statistical
	generality}, in the sense that a classifier fit to some training  dataset,
will work well on another  dataset drawn from the same distribution
to which we do not have access (works well \emph{out-of-sample}).
The \emph{VC dimension }is a key measure of the complexity of a classification
model. The simple $D$-dimensional \emph{linear hyperplane }classification
model, which we discuss in detail below, has VC dimension $D+1$ which
is the lowest of other widely used models such as the decision tree
model (axis-parallel hyper-rectangles, VC dimension $2D$), the\emph{
	$K$}-degree polynomial (VC dimension $O\left(D^{K}\right)$) and
the $L$-layer, $W$-weight piecewise linear deep neural networks
(VC dimension $O\left(WL\log\left(W\right)\right)$), for instance
\citep{vapnik1999nature,blumer1989learnability,bartlett2019nearly}.

Assume a  dataset of size $N$ is drawn i.i.d (independent and identically
distributed) from the same distribution as the training  dataset, according
to \citet{vapnik1999nature}'s \emph{generalization bound theorem},
for the hyperplane classifier we have, with high probability,

\begin{equation}
	E_{\text{test}}\text{\ensuremath{\leq}}E_{\text{emp}}+O\left(\sqrt{\frac{\log\left(N/\left(D+1\right)\right)}{N/\left(D+1\right)}}\right),\label{eq:VC-dimension}
\end{equation}
where $E_{\text{test}}$, $E_{\text{emp}}$ are the \emph{test 0-1
	loss} and \emph{empirical 0-1 loss} of on training dataset, respectively
\citep{mohri2018foundations}. Equation (\ref{eq:VC-dimension}) motivates
finding the exact (gloablly optimal)\emph{ }0-1 loss on the training
data and simplest model, as the lower the training accuracy and the
model complexity (defined by VC-dimension) the more likely the model
will obtain a better result on testing  dataset. If a data set is simple enough, a linear classifier can deliver an accurate enough solution. In which case, no other model can outperform the exact linear classifier.

Training a model to global optimality on a training  dataset is known
as the empirical risk minimization problem. However, even for perhaps
the simplest case---the linear model---training a model to global
optimality is intractable. It has long been proven that empirical
risk minimization for 0-1 loss (i.e., minimizing the number of misclassifications)
in linear classification is NP-hard \citep{ben2003difficulty} as
a function of the data dimension \citep{mohri2018foundations}.

Consequently, most algorithms proposed for this problem focus on optimizing
approximate variants of the 0-1 loss, such as the logistic loss \citep{cox1958regression,cox1966some},
and hinge loss \citep{cortes1995support}. By contrast, relatively
little attention has been given to exact algorithms for the 0-1 loss
classification problem (0-1 LCP). One approach is to formulate the
problem as a mixed-integer program (MIP) and solve it using general-purpose
solvers, such as Gurobi \citep{gurobi}. For instance, \citet{tang2014mixed}
employed a MIP formulation to obtain the maximum-margin boundary under
0-1 loss, while \citet{brooks2011support} optimized the ``ramp loss''
and the hard-margin loss---both closely related to 0-1 loss---using
a quadratic mixed-integer program (QMIP).

Alternatively, combinatorial methods such as the branch-and-bound
(BnB) approach have also been applied. \citet{SIAM-v28-nguyen13a}
for example, proposed several BnB-based algorithms for solving the
0-1 LCP. However, a common problem in BnB research is the lack of
formal proofs of exhaustiveness, making the correctness of such algorithms
uncertain. Although \citet{SIAM-v28-nguyen13a} present several interesting
methods, none of them are accompanied with a formal correctness proof.

Nevertheless, the well-known Cover\textquoteright s functional counting
theorem \citep{cover1965geometrical} rigorously established that
there are 
\begin{equation}
	\mathrm{Cover}\left(N,D+1\right)=2\sum_{d=0}^{D}\left(\begin{array}{c}
		N-1\\
		d
	\end{array}\right)=O\left(N^{D}\right)\label{eq: Cover's functional counting theorem}
\end{equation}
possible \emph{linear dichotomies} of $N$ points in $\mathbb{R}^{D}$.
This result suggests that, in principle, one could solve the 0-1 LCP exactly by exhaustively enumerating
these partitions. However, Cover\textquoteright s result is purely
combinatorial and does not provide any method for performing this
enumeration.

Interestingly, \citet{SIAM-v28-nguyen13a} observed that selecting
hyperplanes formed by choosing $D$ out of $N$ data samples suffices
to solve the 0-1 loss LCP exactly. This
procedure has a combinatorial complexity of $\left(\begin{array}{c}
	N\\
	D
\end{array}\right)$, which appears to be smaller than the complexity derived from Cover\textquoteright s
analysis. At the same time, in the context of the hyperplane decision
tree problem, \citet{murthy1994system,dunn2018optimal} observed
that all possible linear partitions can be enumerated in $2^{D}\left(\begin{array}{c}
	N\\
	D
\end{array}\right)$, which is larger than the bound implied by Cover\textquoteright s
result. These three distinct combinatorial analyses yield seemingly
inconsistent complexity estimates. This naturally raises the question:
\begin{quote}
	\emph{Which of these analyses is correct for solving the 0-1 loss
		linear classification problem? If all are valid, how are they connected?}
\end{quote}
This paper is dedicated to addressing these questions formally. Our key contributions are as follows:
\begin{itemize}
	\item \textbf{Combinatorial foundations for classification in Euclidean
		space}: We establish the combinatorial and incidence relationships
	between hyperplane arrangements and point configurations in the ordinary
	vector space $\mathbb{R}^{D}$. Unlike the classical treatment in combinatorial
	geometry and oriented matroid theory---which is based on homogeneous
	coordinates---we work directly in inhomogeneous (Euclidean) coordinates\footnote{Informally a dataset in $\mathbb{R}^{D}$ is in inhomogeneous coordinates,
		whereas $\bar{\boldsymbol{x}}=\left(\boldsymbol{x},1\right)$ represents
		the same point in homogeneous coordinates (projective space)} \citep{edelsbrunner1987algorithms,fukuda2016lecture}.
	\item \textbf{A novel 0-1 loss linear classification theorem}: We present
	a new Theorem \ref{thm:0-1 loss linear classification theorem} for
	solving the 0-1 LCP, which rigorously proves why \citet{SIAM-v28-nguyen13a}'s
	prioritized combinatorial search (PCS) algorithm can exactly solve
	the 0-1 LCP. The supporting lemmas of Theorem \ref{eq:0-1-loss-linear-classification-problem}
	reveal deep connections between the three distinct combinatorial analyses
	\citet{cover1965geometrical}, \citet{murthy1994system,dunn2018optimal},
	and \citet{SIAM-v28-nguyen13a}.
	\item \textbf{The first rigorously proven standalone algorithm for 0-1 linear
		classification problem}: By combining Theorem \ref{eq:0-1-loss-linear-classification-problem}
	with the efficient combination generator introduced by \citet{he2025CGs},
	we construct the first rigorously proven, standalone algorithm---one
	that does not rely on general-purpose solvers---for solving the 0-1
	LCP. Empirical results (see Figure \ref{fig:ice-vs-bnb}) show that, for example, when $N=150$ data size with $D=3$, ICE would take \textbf{1.2 seconds} worst-case whereas Nguyen and Sanner (2013)'s BnB would take approximately $10^{10}$ seconds (nearly \textbf{317 years}), clearly demonstrating the superiority of our approach.
	\item \textbf{Extension to polynomial hypersurface classification}: We extend
	our theoretical framework to polynomial hypersurfaces, resulting in
	an optimal algorithm for solving the 0-1 loss hypersurface classification
	problem.
	\item \textbf{Empirical insights on generalization}: Our experiments show
	that solutions with lower training accuracy often generalize better
	to unseen test data. This observation refutes the conventional
	belief that exact algorithms overfit and is consistent with \citet{vapnik1999nature}'s
	generalization bound theorem.
\end{itemize}
The paper is organized as follows. In Section \ref{sec:theory}, we
provide a detailed geometric analysis of the linear classification
problem and develop novel theorems for solving the 0-1 loss linear
and hypersurface classification problems. This result leads to a new
class of algorithms capable of solving these problems exactly. Section
\ref{sec:experiments} presents empirical results comparing the ICE
algorithm with standard approximate methods on real-world  datasets
from the UCI Machine Learning Repository, evaluating both training
accuracy and out-of-sample generalization performance. Finally, Section
\ref{sec:summary}, discusses our contributions and the limitations
of the proposed algorithm, and outlines potential directions for future
research.

\section{Theory \label{sec:theory}}

\subsection{Problem definition}

Assume a dataset consists of $N$ \emph{data points} (or data items)
$\boldsymbol{x}_{n}$, $\forall n\in\left\{ 1,\ldots,N\right\} =\mathcal{N}$,
where the data points $\boldsymbol{x}_{n}\in\mathbb{R}^{D}$ and $D$
is the dimension of the\emph{ feature space.} Each data point has
a unique true \emph{label }$l_{n}\in\left\{ -1,1\right\} $, $\forall n\in\mathcal{N}$.
All true labels in this dataset are stored in set $\boldsymbol{l}=\left\{ l_{1},l_{2},...,l_{N}\right\} $.
The data points and their labels are packaged together into the  dataset
$\mathcal{D}$, denoted as $\mathcal{D}_{\boldsymbol{l}}$. The 0-1
LCP can be defined as

\begin{equation}
	\hat{\boldsymbol{w}}=\underset{\boldsymbol{w}\in\mathbb{R}^{D+1}}{\textrm{argmin }}E_{\textrm{0-1}}\left(\boldsymbol{w},\mathcal{D}_{\mathbf{l}}\right)=\sum_{n\in\mathcal{N}}\boldsymbol{1}\left[\textrm{sign}\left(\boldsymbol{w}^{T}\bar{\boldsymbol{x}}_{n}\right)\neq l_{n}\right].\label{eq:0-1-loss-linear-classification-problem}
\end{equation}
where $E_{\textrm{0-1}}\left(\boldsymbol{w},\mathcal{D}_{\mathbf{l}}\right)=\sum_{n\in\mathcal{N}}\boldsymbol{1}\left[\textrm{sign}\left(\boldsymbol{w}^{T}\bar{\boldsymbol{x}}_{n}\right)\neq l_{n}\right]$
is the \emph{0-1 loss objective function} which counts the number
of misclassified data points given the parameter $\boldsymbol{w}$,
we denote $E_{\textrm{0-1}}\left(\boldsymbol{w},\mathcal{D}_{\mathbf{l}}\right)$
as $E_{\textrm{0-1}}\left(\boldsymbol{w}\right)$ when $\mathcal{D}_{\mathbf{l}}$
is clear from the context. The supervised classification problem is
solved by computing (\ref{eq:0-1-loss-linear-classification-problem})
which is a sum of 0-1 loss functions $\boldsymbol{1}\left[\;\right]$,
each taking the value 1 if the Boolean argument is true, and 0 if
false. The function $\text{sign}$ returns $+1$ is the argument is
positive, and $-1$ if negative (and zero otherwise). The linear decision
function $\boldsymbol{w}^{T}\bar{\boldsymbol{x}}$ with parameters
$\boldsymbol{w}\in\mathbb{R}^{D+1}$ and $\bar{\boldsymbol{x}}=\left(\boldsymbol{x},1\right)$
is a data point in \emph{homogeneous coordinates}.
Although apparently simple, this is a surprisingly challenging optimization
problem. Considered as a continuous optimization problem, the standard
ML optimization technique, gradient descent, is not applicable (since
the gradients of $E_{\textrm{0-1}}$ with respect to $w$ are zero everywhere
they exist), and the problem is non-convex so there are a potentially
very large number of local minima in which gradient descent can become
trapped. Nevertheless, the finiteness of the  dataset implies that
only a finite number of partitions are possible. In particular, we
are concerned with those partitions that can be induced by hyperplanes---i.e.,
linear dichotomies. The next subsection explains how to identify these
linear dichotomies using a geometric dual transformation, which can
then be applied to solve (\ref{eq:0-1-loss-linear-classification-problem}).

\begin{figure}
	\begin{centering}
		\includegraphics[viewport=40bp 150bp 1250bp 600bp,clip,scale=0.3]{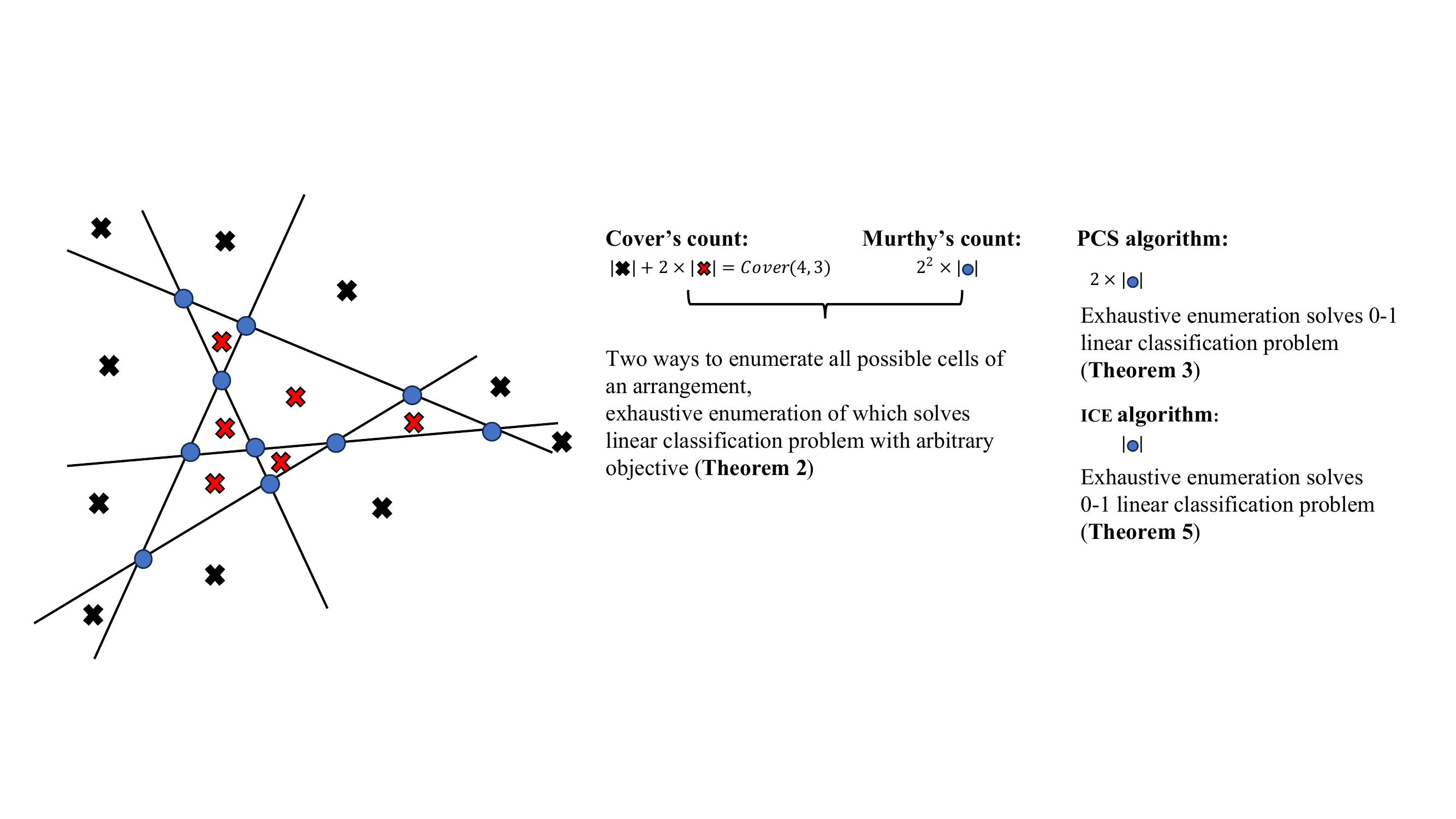}
		\par\end{centering}
	\caption{Novel theoretical contributions enabling the ICE algorithm: identifying the necessary and sufficient dual-arrangement faces that must be enumerated to solve the 0-1 LCP. The black $\boldsymbol{\times}$ marks (unbounded cells) and red $ \color{red}\boldsymbol{\times}$
		marks (bounded cells) represent all the cells of a dual arrangement,
		with  $\left|\cdot\right|$ denoting their size. In Theorem \ref{Linear classification theorem},
		we show that exhaustively enumerating all cells and the reversals
		of unbounded cells (with total size $\left|\boldsymbol{\times}\right|+2\left| {\color{red}\boldsymbol{\times}} \right|$)
		yields a number exactly matching Cover's counting function $\mathit{Cover}$
		for possible linear dichotomies (as proved in Lemma \ref{lem: Cover bound}).
		This procedure solves the linear classification problem for any objective
		function, filling the gap in Cover's theorem, which provides only
		a counting formula without specifying how to enumerate the dichotomies.
		Theorem \ref{thm:0-1 loss linear classification theorem} demonstrates
		that the 0--1 LCP can be solved exactly
		by exhaustively enumerating all \textit{blue circles} in the figure and their
		corresponding reversed sign vectors, formally proving the correctness
		of \citet{SIAM-v28-nguyen13a}'s PCS algorithm, which had only been
		\textbf{empirically observed} to be optimal. Finally, Theorem \ref{Symmetry-fusion-theorem.}
		shows that it suffices to enumerate only the blue circles, without
		their reversed signs, reducing the number of configurations and enabling
		the construction of our \emph{incremental cell enumeration} (ICE)
		algorithm.\label{fig: summary}}
	
\end{figure}

A diagrammatic summary of the key geometric results is presented in Figure \ref{fig: summary}.

\subsection{Point configurations and hyperplane arrangements}

A \emph{point configuration} is synonymous with a  dataset and is denoted
by $\mathcal{P}=\left\{ \boldsymbol{p}_{n}\in\mathbb{R}^{D}:n\in\mathcal{N}\right\} $.
A finite \emph{hyperplane arrangement} is a finite set of hyperplanes
$\mathcal{H}=\left\{ h_{1},...,h_{k}\right\}$, where each hyperplane
is defined as $h_{n}=\left\{ \boldsymbol{x}\in\mathbb{R}^{D}:\boldsymbol{w}^{T}\boldsymbol{x}=c\right\} $
for some constant $c\in\mathbb{R}$. A point configuration or hyperplane
arrangement in \emph{general position} is called \emph{simple} if
no \emph{$k$} of them lie in a \emph{$\left(k-2\right)$-}dimensional
affine subspace of $\mathbb{R}^{D}$ and the intersection of any \emph{k}
hyperplanes is contained in a \emph{$\left(D-k\right)$-}dimensional
\emph{flat},\emph{ }for $1\leq k\leq D$. For example, if $D=2$ then
a set of lines is in general position if no two are parallel and no
three meet at a point.
\begin{defn}
	\emph{Faces of a hyperplane arrangement}. Let $\mathcal{F}_{\mathcal{H}}$
	be the set of all sign vectors $\text{sign}_{\mathcal{H}}\left(\boldsymbol{x}\right)$
	in $\mathbb{R}^{D}$ for arrangement $\mathcal{H}$, which is defined
	as
	\begin{equation}
		\mathcal{F}_{\mathcal{H}}=\left\{ \text{sign}_{\mathcal{H}}\left(\boldsymbol{x}\right):\boldsymbol{x}\in\mathbb{R}^{D}\right\} ,
	\end{equation}
	
	A\emph{ face }$f$ (connected component) of an arrangement $f\subseteq\mathbb{R}^{D}$
	is a maximal subset of $\mathbb{R}^{D}$, such that all $\boldsymbol{x}\in f$
	have the same sign vector $\text{sign}_{\mathcal{H}}\left(\boldsymbol{x}\right)\in\mathcal{F}_{\mathcal{H}}$.
	Given a sign vector $\text{sign}_{\mathcal{H}}\left(\boldsymbol{x}\right)=\left(\delta_{1}\left(\boldsymbol{x}\right),\delta_{2}\left(\boldsymbol{x}\right),\ldots,\delta_{I}\left(\boldsymbol{x}\right)\right)$,
	the connected region of $f$ can be defined as $f=\bigcap_{i\in\mathcal{I}}h_{i}^{\delta_{i}\left(f\right)}$
	. In fact, $f$ defines an \emph{equivalence class }in $\mathbb{R}^{D}$.
	Since any point $\boldsymbol{x}\in f$ has the same sign vector, then
	$\text{sign}_{\mathcal{H}}\left(f\right)$ is \emph{the }sign vector
	for any point in $f$. A face is said to be $k$-\emph{dimensional}
	if it is contained in a $k$-\emph{flat} for $-1\leq k\leq D+1$.
	Some special faces are given specific names \emph{vertices} ($k=0$),\emph{
		edges} ($k=1$), and \emph{cells} ($k=D$). A $k$-face $g$ and a
	$\left(k-1\right)$-face $f$ are said to be \emph{incident }if $f$
	is contained in the boundary\emph{ }of face $g$, for $1\leq k\leq D$.
	In that case, face $g$ is called a \emph{superface} of $f$, and
	$f$ is called a \emph{subface }of $g$. The cells in an arrangement
	can be further split into two classes, the \emph{bounded cells} and\emph{
		unbounded cells}. Informally, a cell is bounded if it is a closed
	region surrounded by hyperplanes (the boundaries are not contained
	in cells), and unbounded otherwise.
\end{defn}
Superficially, a hyperplane arrangement might seem to contain more
information or structure than a set of data points (a point configuration).
However, a valuable approach to studying geometric objects involving
points and hyperplanes is to explore the transformations between these
two objects. By studying the \emph{dual transformation} between point
configurations and hyperplane arrangements, it will later be seen
that the superficial impression of the structural information contained
in hyperplane arrangement and point configuration is incorrect. Both
hyperplane arrangements and point configurations possess equally rich
combinatorial structure.

In the next section, we examine the geometric relationships among
points, hyperplanes, and dichotomies, with a focus on their combinatorial
and incidence relations, leading to a new perspective on the linear
classification problem. This enables the development of
an efficient and general algorithm capable of solving linear classification
problems. \textbf{Detailed proofs of all theorems and lemmas in next section are provided in the Appendix} \ref{appd: Poofs}.

\subsection{Linear classification and point-hyperplane duality}

The geometric dual transformation $\phi:\mathbb{R}^{D}\to\mathbb{R}^{D}$
maps a point $\boldsymbol{p}$ to a non-vertical affine hyperplane
$\phi\left(\boldsymbol{p}\right)$, defined by the equation
\begin{equation}
	p_{1}x_{1}+p_{2}x_{2}+...+p_{D-1}x_{D-1}-x_{D}=p_{D},
\end{equation}
and conversely, the function $\phi^{-1}$ transforms a (non-vertical)
hyperplane $h$ defined by polynomial $w_{1}x_{1}+w_{2}x_{2}+...+w_{D-1}x_{D-1}-x_{D}=w_{D}$
to a point $\phi^{-1}\left(h\right)=\left(w_{1},w_{2},\ldots,w_{D}\right)^{T}$.
The terms\emph{ primal space}, and \emph{dual space} refer to the
spaces before and after transformation by $\phi$ and $\phi^{-1}$.
The dual transformation is naturally extended to a set of points $\phi\left(\mathcal{P}\right)$
and a set of hyperplanes $\phi\left(\mathcal{H}\right)$ by applying
it to all points and hyperplanes in the set. We have the following important
theorem which is the foundation for analysising the incidence and
combinatorial relations between data points and linear dichotomies.
\begin{theorem}
	Incidence relations\index{Incidence relations@\emph{Incidence} \emph{relations}}
	of the dual transformation. \emph{Let $\boldsymbol{p}$ be a point
		and a non-vertical affine hyperplane $h=\left\{ \boldsymbol{x}:\boldsymbol{w}^{T}\boldsymbol{x}=0\right\} $
		in $\mathbb{R}^{D}$. Under the dual transformation $\phi$, $\boldsymbol{p}$
		and $H$ satisfy the following properties:\label{Theorem 6. The Incidence relations of dual transformation}}
	
	\emph{1.} Incidence preservation\emph{: Point $\boldsymbol{p}$ belongs
		to hyperplane $h$ if and only if point $\phi^{-1}\left(h\right)$
		belongs to hyperplane $\phi\left(\boldsymbol{p}\right)=p$,}
	
	\emph{2.} Order preservation\emph{: Point $\boldsymbol{p}$ lies above
		(below) hyperplane $h$ if and only if point $\phi^{-1}\left(h\right)$
		lies above (below) hyperplane $\phi\left(\boldsymbol{p}\right)$.}
\end{theorem}
That the dual transformation preserves the incidence relations above can
be proved by examining the relationship between the dual transformation
$\phi$ and the \emph{unit paraboloid }\citep{edelsbrunner1987algorithms}.
The incidence preservation property described above implies a duality
between the definitions of general position for point configurations
and hyperplane arrangements. For instance, when $D=2$, three points
lying in the same 1-flat $l$ (a line) correspond to three lines in
the dual space intersecting at the same point $\phi\left(l\right)$,
these three lines are mutually parallel if the line $l$ is vertical.

\begin{figure}
	\centering
	\includegraphics[width=0.7\columnwidth]{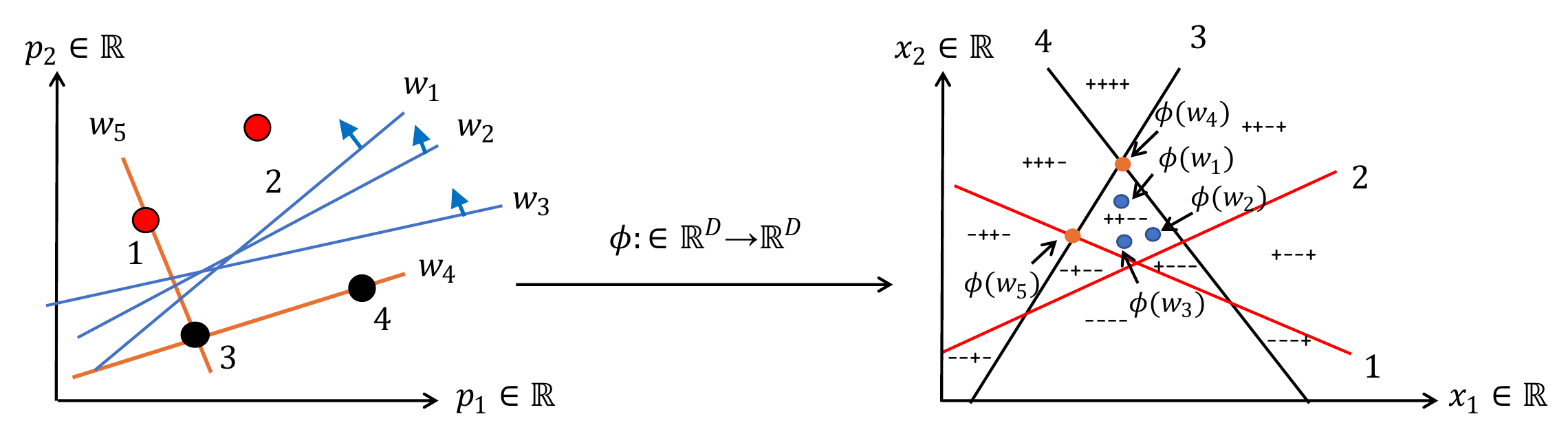}
	
	\caption{A point configuration $\mathcal{D}$ (left-panel) and its dual arrangement
		$\mathcal{H_{D}}$ (right-panel). The yellow hyperplanes $w_{4}$,
		$w_{5}$ with two points lying on them in $\mathbb{R}^{D}$ correspond
		to the yellow points in the dual space, which are the intersection
		of corresponding dual hyperplanes $\phi\left(w_{4}\right)$, $\phi\left(w_{5}\right)$.
		For (blue) hyperplanes $w_{1}$, $w_{2}$, $w_{3}$ with the same
		prediction labels $\left(+,+,-,-\right)$, their corresponding dual
		points $\phi\left(w_{1}\right)$, $\phi\left(w_{2}\right)$, $\phi\left(w_{2}\right)$
		lie in the same cell of dual arrangement $\phi\left(\mathcal{D}\right)$.
		\label{fig:A-point-configuration}}
\end{figure}

It can be difficult to visualize how Cover's dichotomies form equivalence
classes for decision hyperplanes, but the same decision hyperplanes
in the dual space $\phi\left(\mathcal{P}\right)$, $\forall\boldsymbol{p}\in\text{\ensuremath{\mathbb{R}^{D}}}$
partition the space into different cells, where each cell corresponds
to an equivalence class of dichotomies (Fig. \ref{fig:A-point-configuration}).
Moreover, this explains why the prediction by \citet{murthy1994system,dunn2018optimal},
of $2^{D}\left(\begin{array}{c}
	N\\
	D
\end{array}\right)$ possible linear classifications is correct: the most straightforward
way to enumerate all cells in a hyperplane arrangement is to first
enumerate the $\left(\begin{array}{c}
	N\\
	D
\end{array}\right)$ vertices---each determined by a unique combination of $D$ hyperplanes
in general position---and then consider the $2^{D}$ adjacent cells
associated with each vertex. This enumeration relies on the general
position assumption, which guarantees that every $D$-combination of
hyperplanes defines a distinct vertex with exactly $2^{D}$ neighboring
regions.

Importantly, we present the following lemma, which explains the combinatorial
relationship between linear dichotomies and the cells of the dual
arrangement. This lemma is the basis for an alternative approach to proving
Cover\textquoteright s counting theorem.
\begin{lemma}
	\emph{For a set points $\mathcal{D}=\left\{ \mathbf{x}_{n}\in\mathbb{R}^{D}:n\in\mathcal{N}\right\} $
		in general position, the total number of linear dichotomies in Cover's
		function counting theorem, is the same as the number of cells of the
		dual arrangement $\mathcal{H_{D}}$, plus the number of bounded cells
		of $\mathcal{H_{D}}$. In other words, denote the number of dichotomies
		for $N$ data items in $\mathbb{R}^{D}$ as $\mathrm{Cover}\left(N,D+1\right)$
		($D+1$ denote the dimension of data in homogeneous coordinates) and
		the number of cells and bounded cells of an hyperplane arrangement
		in $\mathbb{R}^{D}$ as $B_{D}\left(\mathcal{H_{D}}\right)$ and $C_{D}\left(\mathcal{H_{D}}\right)$.
		Then  \label{lem: Cover bound}
		\begin{equation}
			\mathrm{Cover}\left(N,D+1\right)=B_{D}\left(\mathcal{H_{D}}\right)+C_{D}\left(\mathcal{H_{D}}\right)
		\end{equation}
	}
\end{lemma}
Another, previously reported, geometric
analysis on the combinatorial relations between the hyperplane arrangement
and the point configuration is based on
\emph{homogeneous coordinates}, where all cells of the dual arrangement
are unbounded \citep{edelsbrunner1987algorithms,fukuda2016lecture}.

The equivalence between the number of dichotomies and the sum of the
number of bounded cells and the number of cells may initially seem
unclear. The intuition lies in the fact that not every dichotomy in
the primal space corresponds to a cell in the dual space. Specifically,
decision boundaries associated with unbounded cells correspond to
two dichotomies, whereas those associated with bounded cells correspond
to only one. This relationship is clarified by the following lemma.
\begin{lemma}
	\emph{For a dataset $\mathcal{D}$ in general position, each of Cover's
		dichotomies corresponds to a cell in the dual space, and dichotomies
		corresponding to bounded cells have no }complement cell\emph{ (cells
		with reverse sign vector). Dichotomies corresponding to the unbounded
		cells in the dual arrangements $\phi\left(\mathcal{D}\right)$ have
		a complement cell. \label{Lemma: dichotomies  and cells}}
\end{lemma}
Since each of Cover's dichotomies corresponds to a cell in the dual
space, and dichotomies corresponding to bounded cells have no complement
cell (cells with reverse sign vector), lemma \ref{Lemma: dichotomies  and cells}
demonstrates that all possible\emph{ Cover's dichotomies }of a given
dataset $\mathcal{D}$ can be obtained by enumerating the cells of
an arrangement and the complemented cells of the bounded cells. The
enumeration of the complements of the bounded cells requires an additional
process, as the bounded cells within the arrangement do not have complementary
cells. This result leads directly to the following theorem.
\begin{theorem}
	Linear classification theorem\emph{. Let $\mathcal{D}$ be a data
		set in general position in $\mathbb{R}^{D}$. If an $O\left(N^{D+1}\right)$-time
		cell enumeration algorithm exists, then exact solutions for the linear
		classification problem with an arbitrary objective function can be
		obtained in at most $O\left(t_{\text{eval}}\times N^{D+1}\right)$ time
		by exhaustively enumerating the cells of the dual arrangement $\mathcal{H_{D}}$,
		where $t_{\text{eval}}$ represents the time required to evaluate
		the classification objective. \label{Linear classification theorem}}
\end{theorem}
Theorem \ref{Linear classification theorem} gives us a method for
solving the linear classification problem over \emph{arbitrary} objective
function. However, as we interested in only the LCP with 0-1 loss
objective (\ref{eq:0-1-loss-linear-classification-problem}), the
properties below helps us to solve the LCP over 0-1 loss more efficiently.
The next lemma explains not only that Cover's dichotomies have corresponding
dual cells for the dual hyperplane arrangement, but also that hyperplanes
containing $0\leq k\leq D$ data points have corresponding dual faces.
\begin{lemma}
	\emph{For a dataset $\mathcal{D}$ in general position, a hyperplane
		with $k$ data items lying on it, $0\leq k\leq D$ correspond to a
		$\left(D-k\right)$-face in the dual arrangement $\mathcal{H_{D}}$.
		Hyperplanes with $D$ points lying on it, correspond to vertices in
		the dual arrangement.\label{vertex-hyperplane-connection}}
\end{lemma}
\begin{defn}
	Given a hyperplane arrangement $\mathcal{H}=\left\{ h_{n}:n\in\mathcal{N}\right\} $.
	The separation set $sep\left(f,g\right)$ for two faces $f$, $g$
	is defined by
	\begin{equation}
		\mathrm{sep}\left(f,g\right)=\left\{ n\in\mathcal{N}:\delta_{n}\left(f\right)=-\delta_{n}\left(g\right)\neq0\right\} ,
	\end{equation}
	using which, we say that the two faces $f$, $g$ are \emph{conformal}
	if $\mathrm{sep}\left(f,g\right)=\emptyset$.
	
	That two faces that are conformal is essentially the same thing as
	saying that two faces have consistent classification assignments.
\end{defn}
\begin{lemma}
	\emph{Given a hyperplane arrangement $\mathcal{H}=\left\{ h_{n}:n\in\mathcal{N}\right\} $,
		two faces $f$, $g$ are conformal if and only if $f$ and $g$ are
		subfaces of a common} \emph{face or one face is a subface of the other.}
\end{lemma}
A similar result is described in \emph{oriented matroid }theory \citep{bjorner1999oriented}.
The following lemma will be instrumental in the analysis, presented
later, of the linear classification problem with the \emph{0-1 loss
}objective. It suggests that the optimal cell, with respect to 0-1
loss, is conformal to the optimal vertex.
\begin{lemma}
	\emph{Given a hyperplane arrangement $\mathcal{H}=\left\{ h_{n}:n\in\mathcal{N}\right\} $,
		for an arbitrary maximal face (cell) $f$, the sign vector of $f$
		is $\text{sign}_{\mathcal{H}}\left(f\right)$. For an arbitrary $\left(D-d\right)$-dimension
		face $g$, $0<d\leq D$, the number of different signs of $\text{sign}_{\mathcal{H}}\left(g\right)$
		with respect to $\text{sign}_{\mathcal{H}}\left(f\right)$ is larger
		than or equal to $d$, where equality holds only when $g$ is conformal
		to $f$ ($g$ is a subface of $f$). \label{optimal conformal faces}}
\end{lemma}
Now we have all receipts to prove the final result, for the linear
classification problem over 0-1 loss, we can solve it by exhuastively
searching all $D$-combinations of data points. The following theorem
formally proves \citet{SIAM-v28-nguyen13a}'s observation.
\begin{theorem}
	0-1 loss linear classification theorem\emph{. Consider a  dataset $\mathcal{D}_{\mathbf{l}}$
		of $N$ data points of dimension $D$ in general position, along with
		their associated labels. Let $\mathcal{S}_{\text{kcombs}}$ denote
		the set of all $D$-combinations with respect to  dataset $\mathcal{D}$.
		Then we have following relation
\begin{equation}
	\underset{s\in\mathcal{S}_{\text{kcombs}}\left(D,\mathcal{D}\right)}{\text{argmin}}
	\min\!\left(E_{\text{0-1}}\!\left(\boldsymbol{w}_{s},\mathcal{D}_{\mathbf{l}}\right),
	E_{\text{0-1}}\!\left(-\boldsymbol{w}_{s},\mathcal{D}_{\mathbf{l}}\right)\right)
	\subseteq
	\underset{\boldsymbol{w}\in\mathbb{R}^{D+1}}{\text{argmin}}
	E_{\text{0-1}}\!\left(\boldsymbol{w},\mathcal{D}_{\mathbf{l}}\right)
	\label{eq:0-1-loss-classification-theorem}
\end{equation}
		where $\boldsymbol{w}_{s}$ represents the normal vector of the hyperplane
		that pass through the $D$-combination of data $s$, and $-\boldsymbol{w}_{s}$
		is the negation of $\boldsymbol{w}_{s}$. The inner $\min$ on the
		left-hand side ensures that $s\in\mathcal{S}_{\text{kcombs}}\left(D,\mathcal{D}\right)$ for each $s$
		, where $\mathcal{S}_{\text{kcombs}}\left(D,\mathcal{D}\right)$ denote
		all possible $D$-combinations of the set $\mathcal{D}$. We take
		the smaller of $E_{\text{0-1}}\left(\boldsymbol{w}_{s}\right)$ and
		$E_{\text{0-1}}\left(-\boldsymbol{w}_{s}\right)$, and the outer $\text{argmin}$
		selects} one\emph{ of the values of that minimizes this quantity over
		all $s\in\mathcal{S}_{\text{kcombs}}\left(D,\mathcal{D}\right)$. \label{thm:0-1 loss linear classification theorem}}
\end{theorem}

\subsection{Non-linear (polynomial hypersurface) classification}

Based on the point-hyperplane duality, equivalence relations for linear
classifiers on finite sets of data were established above. However,
a linear classifier is often too restrictive in practice, as many
problems require more complex decision boundaries. It is natural to
ask whether it is possible to extend the theory to non-linear classification.
This section examines a well-known concept in algebraic geometry, the
\emph{$K$-tuple Veronese embedding}\index{Veronese embedding}, which
allows the generalization of the previous strategy for solving classification
problem with \emph{hyperplane classifier }to problems involving \emph{hypersurface
	classifiers}.

Importantly, we present the following theorem, which describes the
relationship between hyperplane and hypersurface classification problems.

\begin{theorem}
	The $K$-tuple Veronese embedding. \emph{Given variables $x_{0},x_{1},\ldots x_{D}$
		in projective space $\mathbb{P}^{D}$ (which is isomorphic to the
		affine space $\mathbb{R}^{D}$ when ignoring the points at infinity
		\citep{cox1997ideals}), let $M_{0},M_{1},\ldots M_{G}$ be all monomials
		of degree $K$ with variables $x_{0},x_{1},\ldots x_{D}$, where $G=\left(\begin{array}{c}
			D+K\\
			D
		\end{array}\right)-1$ (see Appendix \ref{appd: Poofs} for the formal definition of monomials
		and polynomials and explanation of $G$). Define a mapping $\rho_{K}:\mathbb{P}^{D}\to\mathbb{P}^{G}$
		which sends the point $\bar{\boldsymbol{p}}=\left(p_{0},p_{1},\ldots p_{D}\right)\in\mathbb{P}^{D}$
		to the point $\rho_{K}\left(\bar{\boldsymbol{p}}\right)=\left(M_{0}\left(\bar{\boldsymbol{p}}\right),M_{1}\left(\bar{\boldsymbol{p}}\right),\ldots M_{G}\left(\bar{\boldsymbol{p}}\right)\right)$.
		This is called the $K$-tuple Veronese embedding of $\mathbb{P}^{D}$
		in $\mathbb{P}^{G}$. The hyperplane classification over the embedded
		 datasets $\rho_{K}\left(\mathcal{D}\right)$ is isomorphic to the
		polynomial hypersurface classification (defined by a degree $K$ polynomial)
		over the original  dataset $\mathcal{D}$.}
\end{theorem}

It is now straightforward to extend Theorem \ref{thm:0-1 loss linear classification theorem}
to the following polynomial hypersurface classification theorem.
\begin{cor}
	0-1 loss polynomial hypersurface classification theorem. \emph{Consider
		a  dataset $xs$ of $N$ data points in $\mathbb{R}^{D}$ in general
		position, along with their associated labels. Let $\rho_{K}\left(\mathcal{D}\right)$
		be the $K$-tuple Veronese embedding defined by monomials of degree
		$K$, we have following relation}
	
\begin{equation}
	\underset{s\in\mathcal{S}_{\text{kcombs}}\left(G,\rho_{K}\left(\mathcal{D}\right)\right)}{\text{argmin}}
	\min\!\left(E_{\text{0-1}}\!\left(\boldsymbol{w}_{s},\rho_{K}\left(\mathcal{D}_{\mathbf{l}}\right)\right),
	E_{\text{0-1}}\!\left(-\boldsymbol{w}_{s},\rho_{K}\left(\mathcal{D}_{\mathbf{l}}\right)\right)\right)
	\subseteq
	\underset{\boldsymbol{w}\in\mathbb{R}^{G+1}}{\text{argmin}}
	E_{\text{0-1}}\!\left(\boldsymbol{w},\rho_{K}\left(\mathcal{D}_{\mathbf{l}}\right)\right)
\end{equation}
	\emph{where $\boldsymbol{w}_{s}\in\mathbb{R}^{G}$ denote as the normal
		vector determined by $s$ ($G$ data points).}
	\label{cor:0-1-loss hypersurface classify}
\end{cor}

\subsection{Incremental cell enumeration (ICE) algorithm\label{subsec:Incremental-cell-enumeration}}

\begin{algorithm}[tb]
	\tiny
	\caption{Incremental cell enumeration (ICE) algorithm}
	\label{alg:algorithm}
	\textbf{Input}: $\mathcal{D}$: input  dataset which consists of $N$
	data points in $\mathbb{R}^{D}$ in general position; $\mathbf{l}$:
	label vector; $K$: degree of the polynomial;\\
	\textbf{Output}: The optimal normal vector $\boldsymbol{w}^*:\mathbb{R}^{D+1}$
	and optimal 0-1 loss $E_{\text{0-1}}^{*}$
	\begin{algorithmic}[1] 
		\STATE $\mathcal{D}^{\prime}=\rho_{K}\left(\mathcal{D}\right)$ // \textit{calculating embedded  datasets}
		\STATE $\boldsymbol{w}^{*}\gets \mathit{svm}\left(\mathcal{D}_{\mathbf{l}}^{\prime}\right)$ 
	    \STATE $\mathit{ds} \gets \mathit{reorder}(\boldsymbol{w}^*, \mathcal{D}')$ \textit{// sort by $|\boldsymbol{w}^\top \boldsymbol{x}|$}
		\STATE $\mathit{Css} \gets [\:[\ ], [\ ], \dots, [\ ]\:]$ \textit{// $K{+}1$ empty lists for 0 to $K$-combinations}

		\FOR{$n = 0$ \textbf{to} $N-1$}
		\FOR{$k= \min{\left(K,n+1\right)}$ \textbf{down to} $0$}
		\STATE $\mathit{Css}[k] \gets \mathit{Css}[k] \cup \mathit{map}(\lambda S.\ S  \mathbin{+\!\!\!+} \left[n\right],\ \mathit{Css}[k-1])$ \textit{// incremental combination generation}
		\ENDFOR
		
		\STATE $\boldsymbol{ws} \gets \mathit{map}(\mathit{genModel}(\mathit{ds}),\ \mathit{Css}[D])$ \textit{// generate normal vectors from combinations}
		
		\FORALL{$\boldsymbol{w}' \in \boldsymbol{ws}$}
		\IF{$E_{\text{0-1}}(\boldsymbol{w}') \leq E_{\text{0-1}}(\boldsymbol{w}^*)$}
		\STATE $\boldsymbol{w}^*, E_{\text{0-1}}^* \gets \boldsymbol{w}', E_{\text{0-1}}(\boldsymbol{w}')$
		\ENDIF
		\IF{$N - D - E_{\text{0-1}}(\boldsymbol{w}') \leq E_{\text{0-1}}(\boldsymbol{w}^*)$} 
		\STATE $\boldsymbol{w}^*, E_{\text{0-1}}^* \gets -\boldsymbol{w}', N - D - E_{\text{0-1}}(\boldsymbol{w}')$ 
		\textit{// symmetric fusion law}
		\ENDIF
		\ENDFOR
		
		\STATE $\mathit{Css}[D] \gets [\ ]$ \textit{// eliminate $D$-combinations after use}
		\ENDFOR
		
		\STATE \textbf{return} $\boldsymbol{w}^*, E_{\text{0-1}}^*$
	\end{algorithmic}\label{alg:Incremental-cell-enumeration}
\end{algorithm}

Due to the \emph{symmetry} of the 0-1 loss, where a data item is assigned
a label of either $1$ or $-1$, the 0-1 loss for the negative orientation
of a hyperplane can be directly derived from the positive orientation
of the same hyperplane without calculating it explicitly. The following theorem formalizes this
relationship.
\begin{theorem}
	Symmetry fusion theorem\emph{. Consider a  dataset $\mathcal{D}$ of
		$N$ data points of dimension $D$ in general position, along with
		their associated labels. Let $h$ be a hyperplane which goes through
		$D$ out of $N$ data points in the  dataset $\mathcal{D}$, separating
		the  dataset into two disjoint sets $\mathcal{D}^{+}$ and $\mathcal{D}^{-}$.
		If the 0-1 loss for the positive orientation of this hyperplane is
		$l$, then the 0-1 loss for the negative orientation of this hyperplane
		is $N-l-D$. \label{Symmetry-fusion-theorem.}}
\end{theorem}
Therefore, the 0-1 loss
linear classification problem can be solved by enumerating only the
positive or negative-oriented hyperplanes, rather than both.

We now have all the necessary components to construct our algorithm,
which enumerates all linear classification decision hyperplanes and
thus solves (\ref{eq:0-1-loss-linear-classification-problem}). Theorem
\ref{thm:0-1 loss linear classification theorem} states that all
globally optimal solutions to this problem are equivalent (in terms
of 0-1 loss) to the optimal solutions contained within the set of
positive and negatively oriented linear classification decision hyperplanes
(vertices in the dual space) passing through $D$ out of $N$ data
points in the  dataset $\mathcal{D}$. There exist numerous algorithms
for enumerating combinations; for example, \citet{SIAM-v28-nguyen13a}'s
PCS algorithm employed a one-by-one enumeration strategy. However,
such a one-by-one approach is inefficient and unsuitable for optimization
tasks, as it is non-recursive and therefore precludes the use of bounding
methods for further acceleration. 

\citet{he2025ROF} and \citet{he2025CGs} provide an extensive discussion
of various combination generators defined in both sequential and divide-and-conquer
styles. We adopt the sequential generator introduced by \citet{he2025CGs}. The
pseudocode is presented in Algorithm \ref{alg:Incremental-cell-enumeration}.
The algorithm has a complexity of $O\left(N^{G+1}\times G^{3}\right)$,
where $G$ is the dimension of the embedded space (with $G=D$ if
$K=1$). Since in line 18 of Algorithm \ref{alg:Incremental-cell-enumeration}
we eliminate $D$-combinations at every recursive step, the algorithm\textquoteright s
memory usage is $O\left(N^{G}\right)$.

\section{Empirical experiments\label{sec:experiments}}

In this section, we analyze the performance of our ICE algorithm empirically. Our evaluation aims to test the following hypotheses: (a) the ICE algorithm consistently achieves the highest training accuracy among competing algorithms when allowing ICE to run to termination; (b) the solutions with significantly higher training accuracy (learned using the ICE algorithm) also achieve higher accuracy on the test  datasets, and (c) the observed wall-clock runtime aligns with the worst-case time complexity analysis.

\paragraph{Exact linear (hyperplane) classification}
We first compare our exact algorithm, ICE, against support vector machines
(SVM)\footnote{We tuned the SVM hyperparameters using a standard coarse grid search,
	testing a set of widely spaced values (e.g., $\left[0.01,0.1,1,\ldots,10000\right]$)
	on a logarithmic scale.}, logistic regression (LR), and linear discriminant analysis
(LDA) on linear setting, using binary classification  datasets from the UCI machine learning
repository \citep{UCI}. As shown in Table \ref{tab:empirical-error-comparisons},
the ICE algorithm consistently finds solutions with lower 0-1 loss
than approximate algorithms.

\begin{table}[t]
	\centering
	\caption{Comparison of the accuracy of our novel ICE algorithm, against approximate methods on real-world
		 datasets. Best performing algorithm is marked bold. \label{tab:empirical-error-comparisons}\\
		}
	\resizebox{0.5\columnwidth}{!}{
		\begin{tabular}{l@{\hskip 4pt}c@{\hskip 4pt}crrrr}
			
			 dataset & $N$ & $D$ & ICE(\%) & SVM(\%) & LR(\%) & LDA(\%)
			\\ \hline 
			HA & 283 & 3 & \textbf{77.03} & 72.08 & 73.14 & 73.85\\
			
			CA & 72 & 5 & \textbf{80.6} & 77.2 & 73.6 & 75.0\\
			
			CR & 89 & 6 & \textbf{95.51} & 91.10 & 89.89 & 89.89\\
			
			VP & 704 & 2 & \textbf{97.30} & 96.88 & 96.59 & 96.59 \\
			
			BT & 502 & 4 & \textbf{78.69} & 74.50 & 75.50 & 74.10 \\
			
			SP & 975 & 3 & \textbf{94.46} & 94.05 & 94.05 & 94.05 \\
			
	\end{tabular}}
	
\end{table}

Due to space constraints, the results of the runtime complexity analysis
and out-of-sample tests are presented in Appendix \ref{sec:Additional experiments}.
Figure \ref{fig:run-time-polynomial}
shows that the empirical wall-clock runtime agrees closely with the
theoretical predictions. In Figure \ref{fig:ice-vs-bnb},
we compare ICE against the state-of-the-art BnB algorithm by \citet{SIAM-v28-nguyen13a}
for solving the 0-1 LCP. Our empirical
analysis demonstrates that \citet{SIAM-v28-nguyen13a}'s algorithm
exhibits \textit{exponential complexity} in the worst-case. 

Additionally, we also compare the performance of the ICE and BnB algorithms with that of a mixed-integer programming (MIP) solver for the 0–1 LCP, implemented in MATLAB using the GLPK solver, results shown in Figure (Figure \ref{fig: ice-vs-bnb-mip}.). These show that while the MIP solver is more efficient than BnB on small datasets, its performance is less predictable compared with ICE and BnB. 

From (\ref{eq:VC-dimension}), \textit{we anticipate that exact solutions
will not only achieve lower 0-1 loss on training  datasets but are
also more likely to generalize better}, yielding lower 0-1 loss on
test  datasets. To evaluate this hypothesis, Table \ref{tab:Additional Oos}
reports the out-of-sample performance of the ICE algorithm using 5-fold
cross-validation, compared against approximate algorithms. The results indicate that training a linear model with substantially
lower training error than the approximate algorithms also leads to
stronger generalization in out-of-sample tests, thereby refuting the
notion that the optimal solution necessarily overfits the data.

\paragraph{Exact hypersurface (quadratic hypersurface) classification}

\begin{figure}
	\begin{centering}
		\includegraphics[viewport=110bp 70bp 820bp 640bp,clip,scale=0.10]{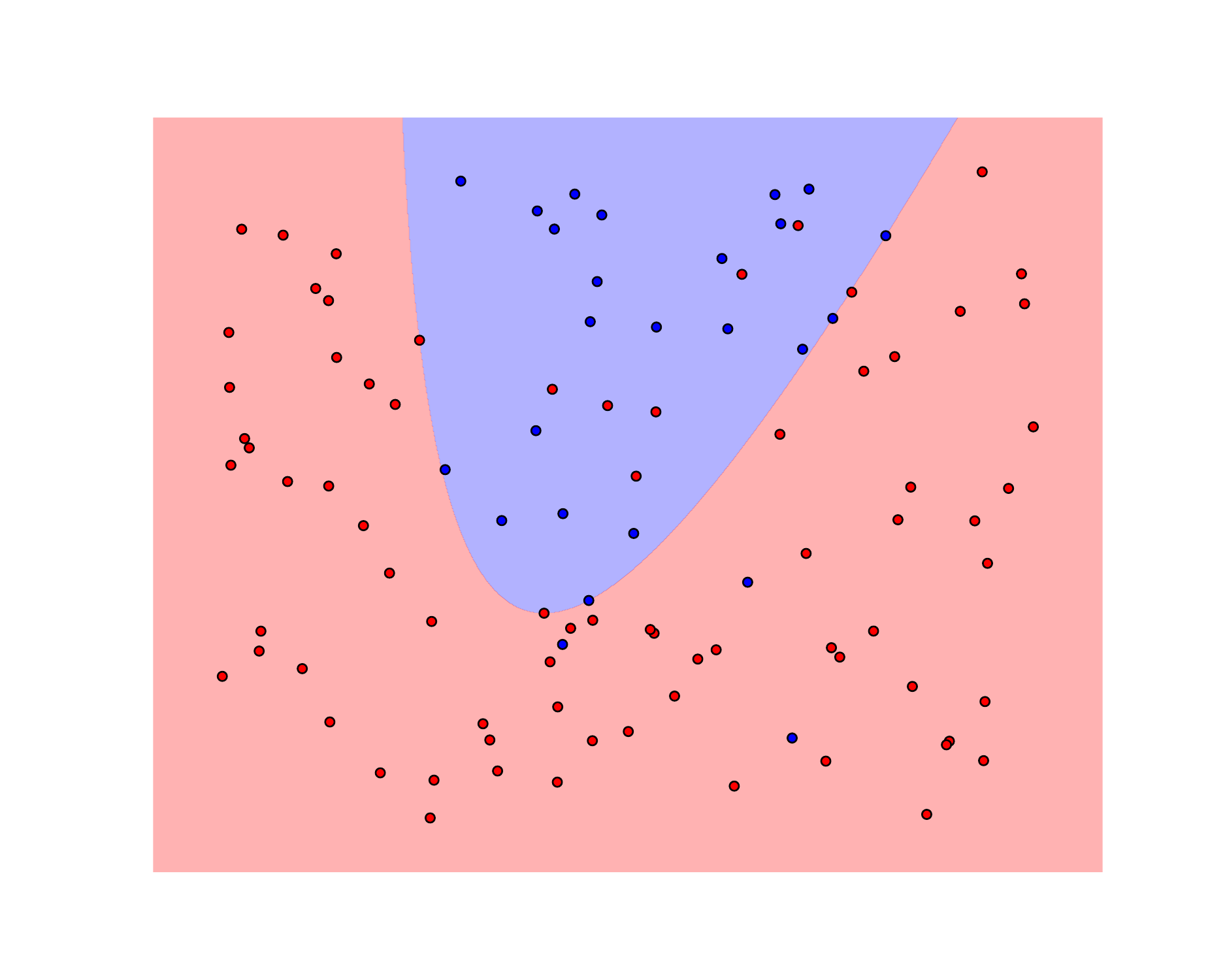}\includegraphics[viewport=110bp 70bp 820bp 640bp,clip,scale=0.10]{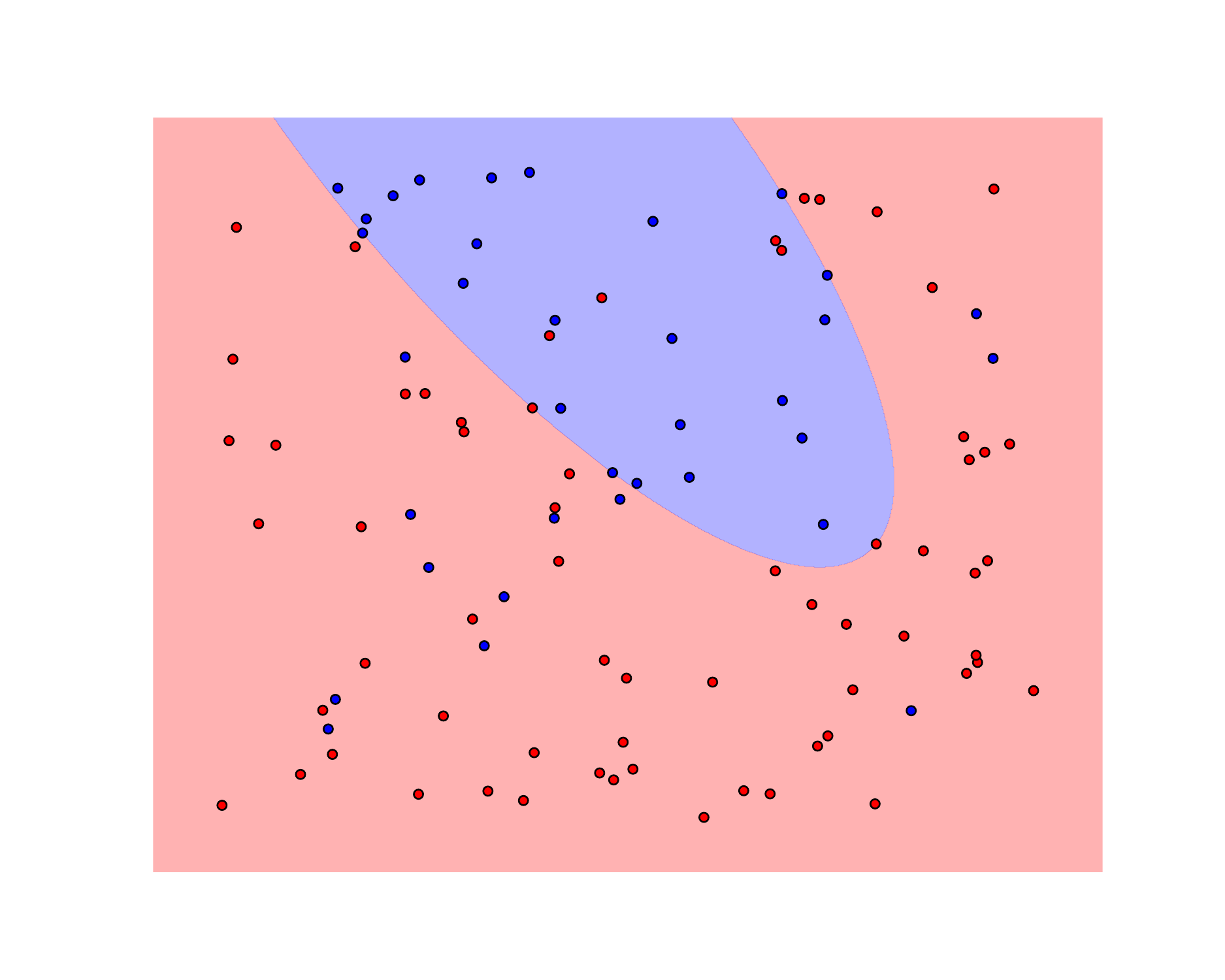}\includegraphics[viewport=110bp 70bp 820bp 640bp,clip,scale=0.10]{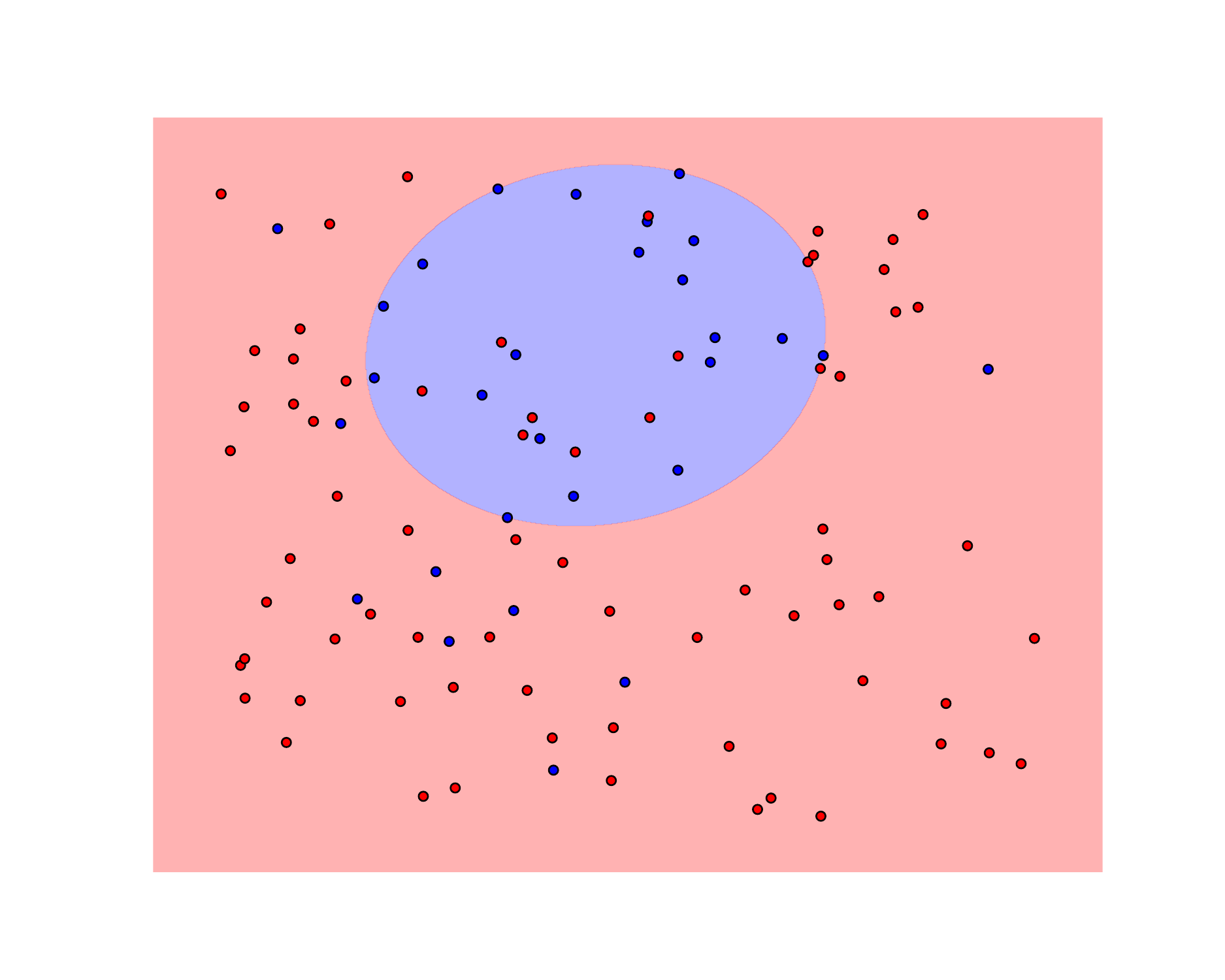}\includegraphics[viewport=110bp 70bp 820bp 640bp,clip,scale=0.10]{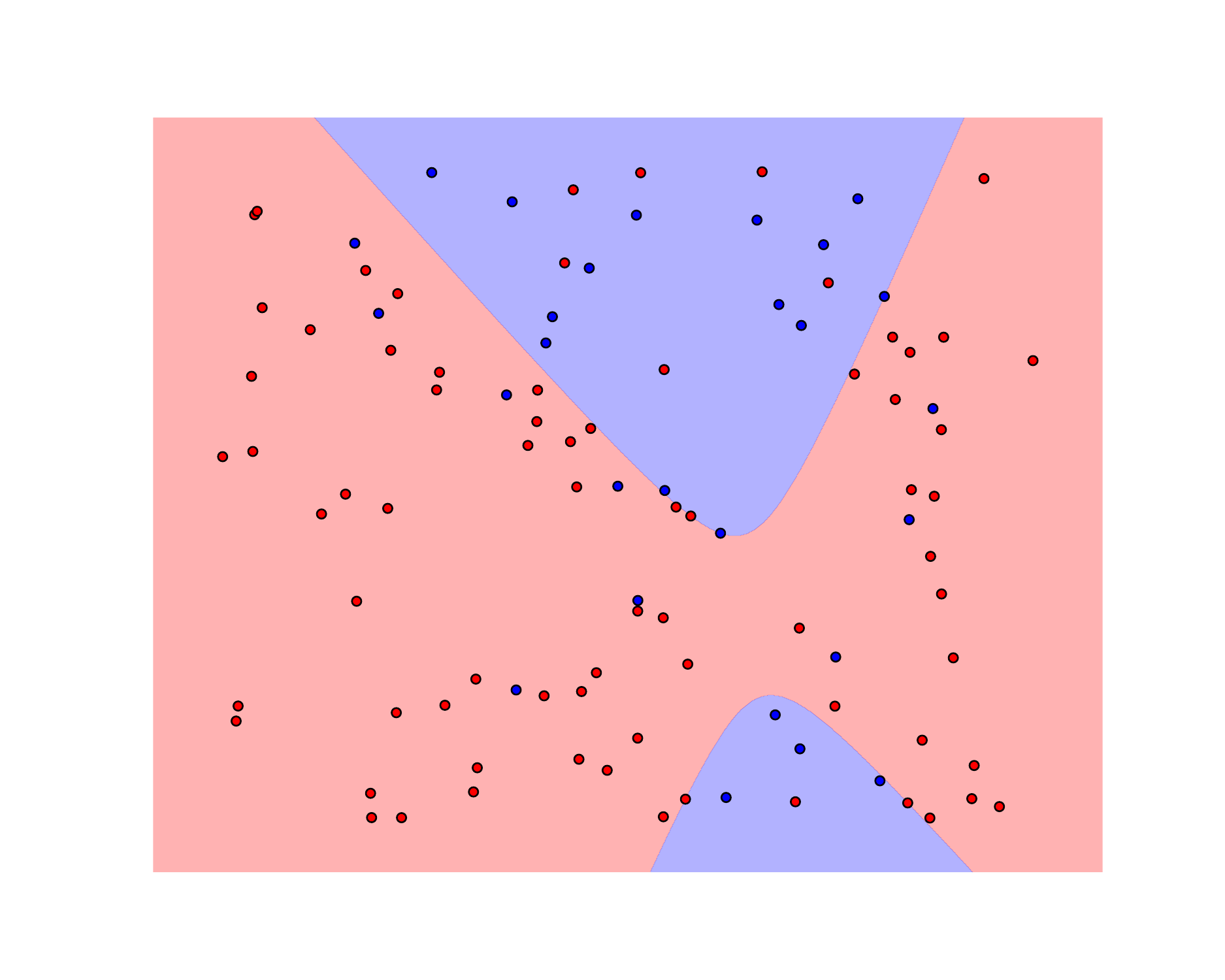}
		\par\end{centering}
	\begin{centering}
		\includegraphics[viewport=110bp 70bp 820bp 640bp,clip,scale=0.10]{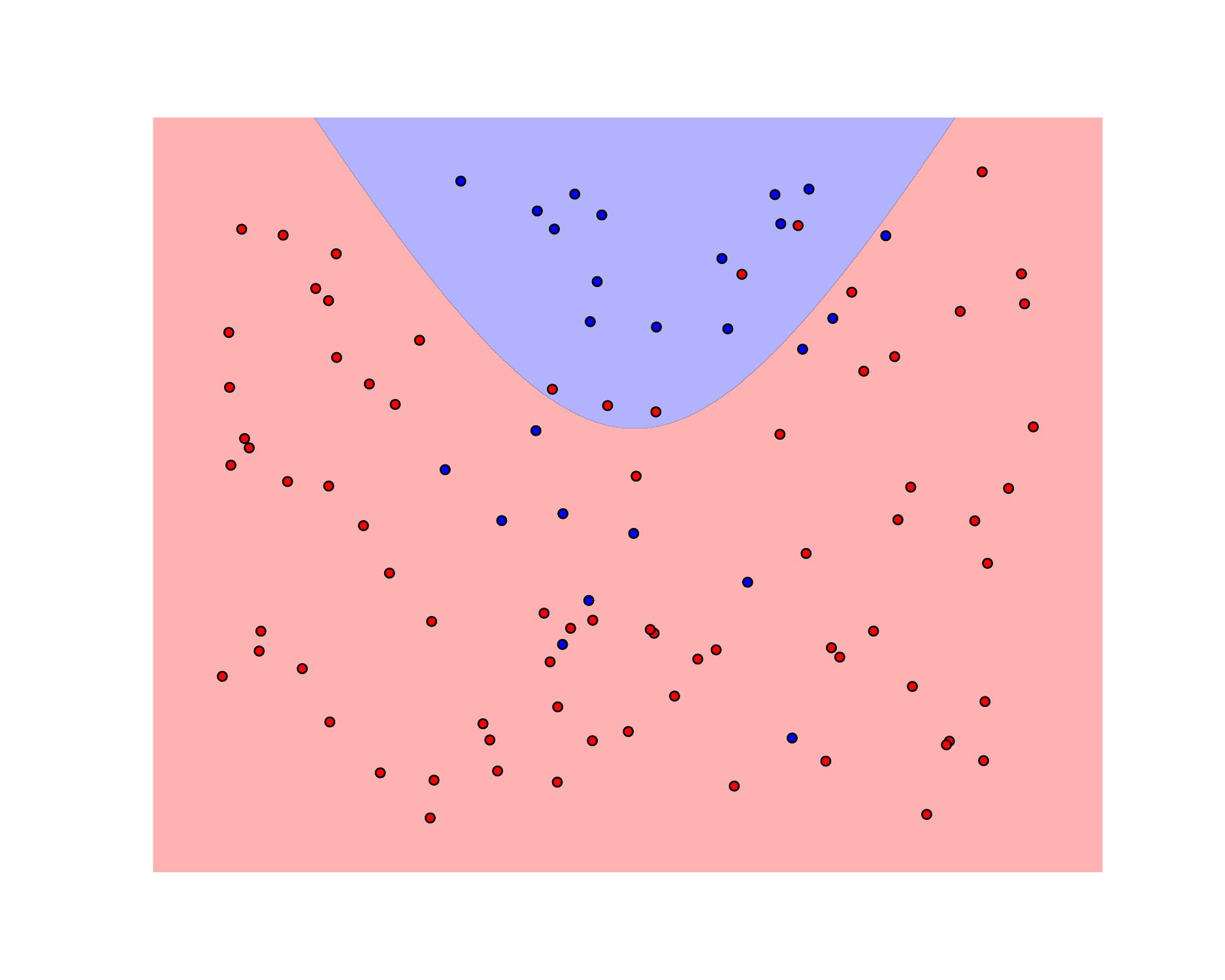}\includegraphics[viewport=110bp 70bp 820bp 640bp,clip,scale=0.10]{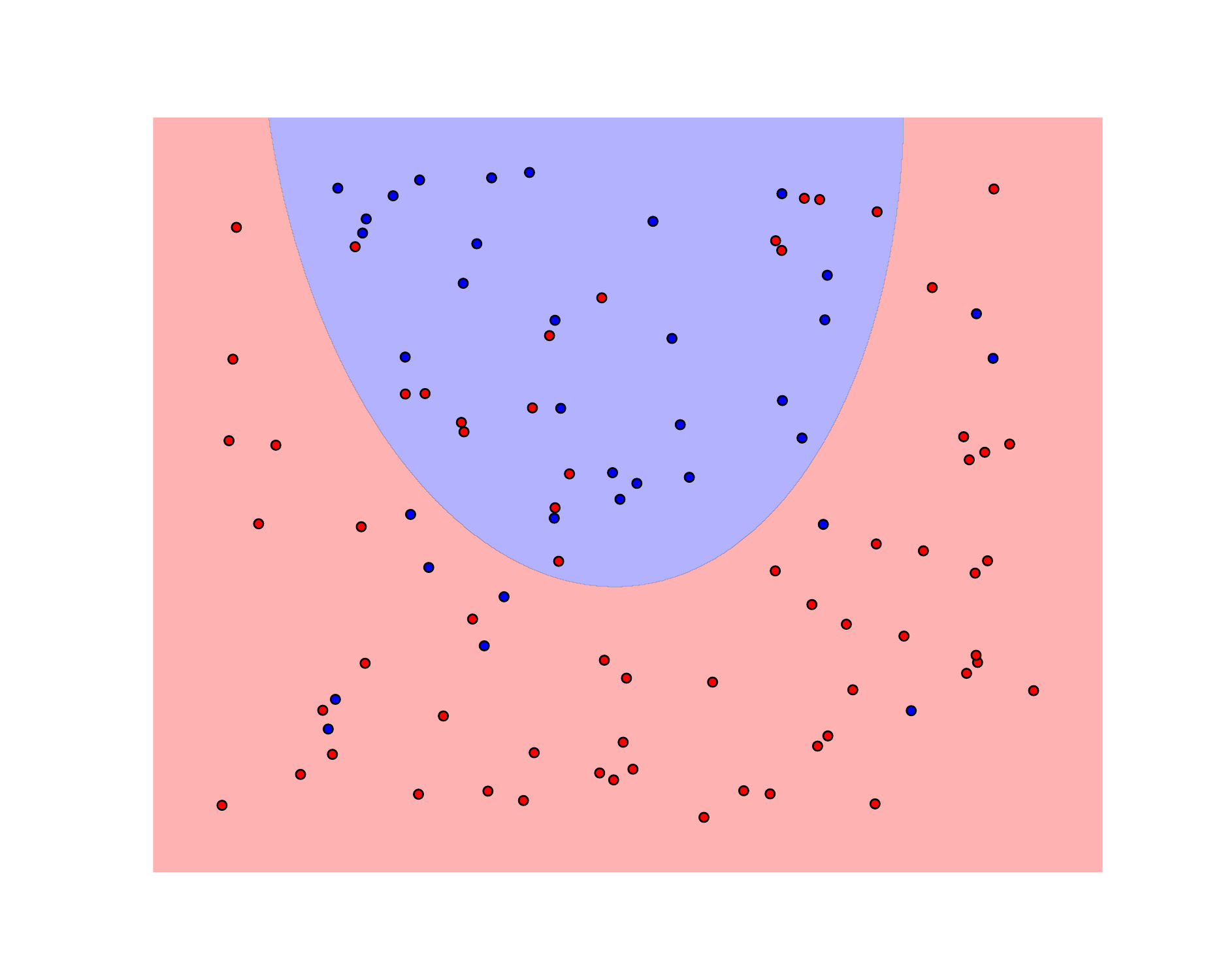}\includegraphics[viewport=110bp 70bp 820bp 640bp,clip,scale=0.10]{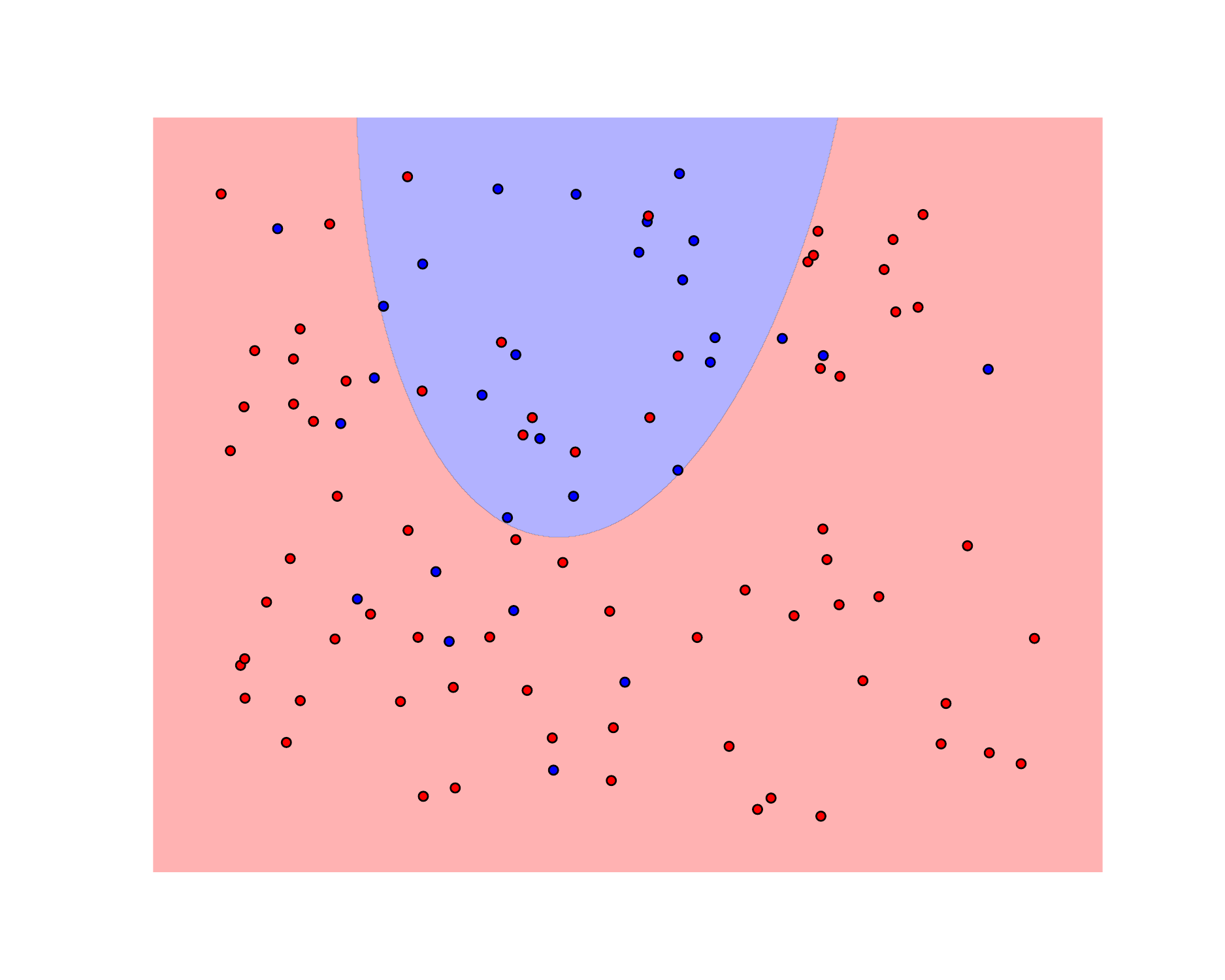}\includegraphics[viewport=110bp 70bp 820bp 640bp,clip,scale=0.10]{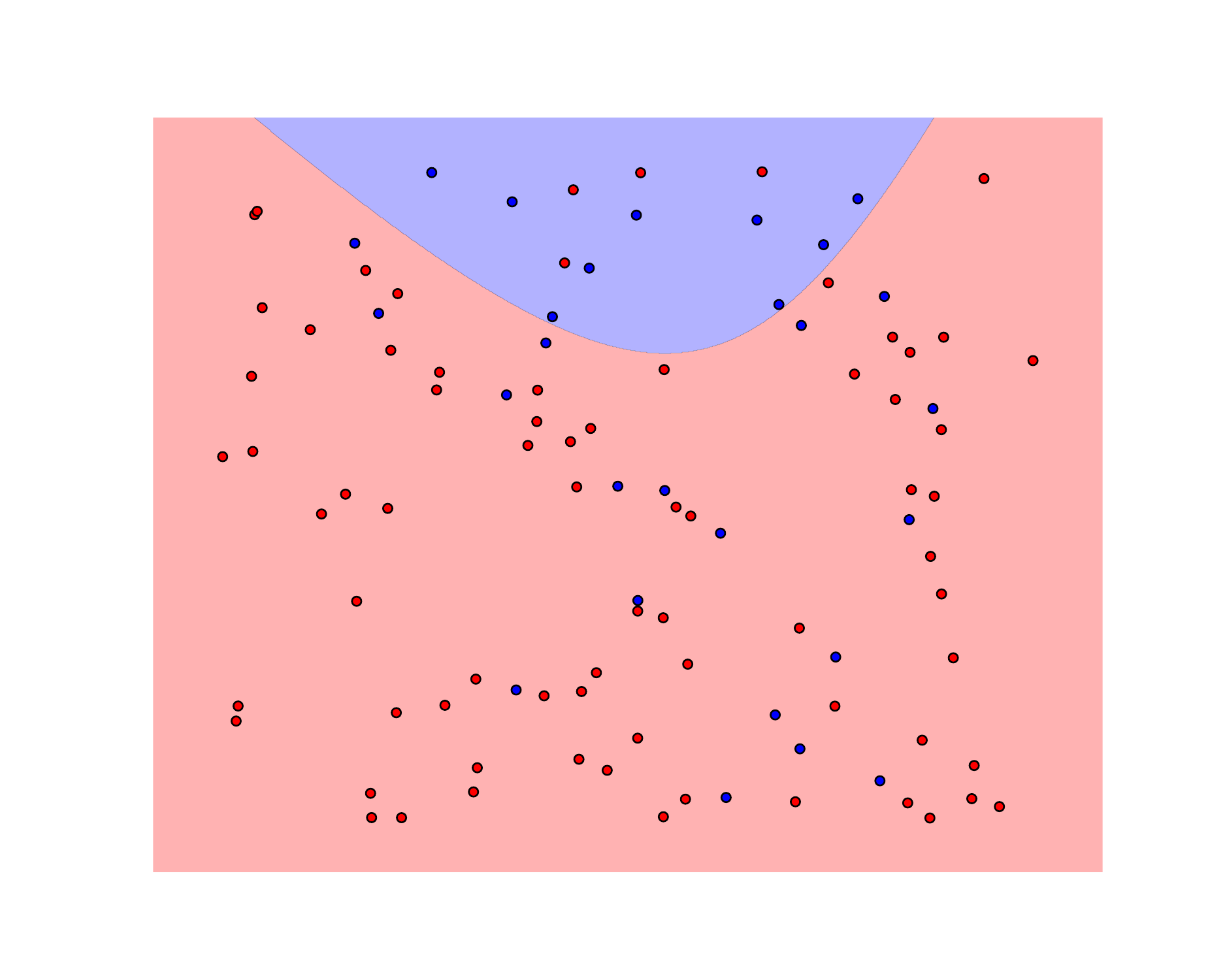}
		\par\end{centering}
	\caption{Optimal quadratic classifiers learned by the ICE algorithm (top four panels) achieve 0–1 losses of 9, 16, 17, and 16, while the approximate quadratic classifiers learned by an SVM with a degree-2 polynomial kernel (bottom four panels) obtain 0–1 losses of 17, 26, 21, and 22.\label{fig:Optimal-quadratic-classifiers}}
\end{figure}

To evaluate the algorithm beyond linear classification, we test the
ICE algorithm on synthetic  datasets whose ground truth is a noisy
quadratic boundary with label noise. We compute the exact solution
(learned by ICE) on four  datasets of size $N=100$ and $D=2$ and
compare it against approximate solutions (learned by SVM with a degree-2
polynomial kernel). The results are shown in Figure \ref{fig:Optimal-quadratic-classifiers}.
Similarly, the out-of-sample generalization performance on real-world
 datasets for the quadratic classifier is reported in Table \ref{tab:quadratic compare}.
	
	\section{Summary, discussion and future work\label{sec:summary}}
	
	In this paper, we have presented \emph{incremental cell enumeration},
	ICE, the first provably correct, worst-case polynomial $O\left(N^{D+1}\right)$, which is polynomial in $N$ and exponential in varying $D$,
	run-time complexity algorithm for solving the 0-1 loss linear classification
	problem (\ref{eq:0-1-loss-linear-classification-problem}). Our empirical
	investigations show that the exact solution often significantly outperforms
	the best approximate solutions on the training  dataset and also yields
	lower test error. This finding is critically important because it
	demonstrates that, contrary to widely held belief, globally optimal solutions to the 0-1 LCP can generalize well to unseen data. Prior to the development of ICE,
	provably correct exact algorithms---such as those proposed by \citet{SIAM-v28-nguyen13a}---were
	computationally intractable even for moderate $N$ and small $D$,
	and their optimality had not been rigorously proved.
	
	The immediate shortcoming of the algorithm is its exponential complexity
	in the data dimension $D$. This combinatorial complexity is further
	compounded in the hypersurface case, where the embedding space has
	dimension $O\left(D^{K}\right)$, resulting in a final hypersurface classification algorithm with
	time complexity $\left(N^{D^{K}}\right)$. However, since this problem
	is NP-hard, the exponential dependence on $D$ and $K$ is unlikely
	to be eliminated unless NP=P. Notably, the ICE algorithm relies solely
	on matrix operations, allowing for full vectorization and parallelization.
	Our current implementation uses simple parallelization via the PyTorch
	library. More sophisticated parallel implementations can be achieved
	by adopting the divide-and-conquer (D\&C) combination generator introduced
	in \citep{he2025CGs}. A parallel implementation based on D\&C-style
	recursion, executed on massively parallel GPUs, is expected to yield significantly better performance.
	
	
\bibliography{iclr2026_conference}
\bibliographystyle{iclr2026_conference}

	\appendix
	
	\section{Proofs and definitions \label{appd: Poofs}}
	\begin{defn}
		\emph{Monomial}. A \emph{monomial\index{monomial}} with respect to
		a $D$-tuple $\boldsymbol{x}=\left(x_{1},x_{2}\ldots,x_{D}\right)$
		is a product of the form 
		\begin{equation}
			M=\boldsymbol{x}^{\boldsymbol{\alpha}}=x_{1}^{\alpha_{1}}\cdot x_{2}^{\alpha_{2}}\ldots\cdot x_{D}^{\alpha_{D}},
		\end{equation}
		where $\boldsymbol{\alpha}=\left(\alpha_{1},\alpha_{2}\ldots,\alpha_{D}\right)$
		and $\alpha_{1},\alpha_{2}\ldots,\alpha_{D}$ are nonnegative integers.
		The\emph{ total degree} of this monomial is the sum $\left|\boldsymbol{\alpha}\right|=\alpha_{1}+\cdot\cdot\cdot+\alpha_{n}$.
		When $\boldsymbol{\alpha}=\boldsymbol{0}=\left(0,\ldots,0\right)$,
		$\boldsymbol{x}^{\boldsymbol{0}}=1$.
	\end{defn}
	\begin{defn}
		\emph{Polynomial}. A polynomial $P$ in $x_{1},x_{2}\ldots,x_{D}$
		with coefficients in $\mathbb{R}$ is a finite linear combination
		(with coefficients in the field $\mathbb{R}$) of monomials. A polynomial
		$P\left(\boldsymbol{x}\right)$, or $P$ in short, will be given in
		the form
		\begin{equation}
			P\left(\boldsymbol{x}\right)=\sum_{i}w_{i}\boldsymbol{x}^{\boldsymbol{\alpha}_{i}},w_{i}\in\mathbb{R},
		\end{equation}
		where $i$ is finite. The set of all polynomials with variables $x_{1},x_{2}\ldots,x_{D}$
		and coefficients in $\mathbb{R}$ is denoted by $\mathbb{R}\left[x_{1},x_{2}\ldots,x_{D}\right]$
		or $\mathbb{R}\left[\boldsymbol{x}\right]$.
	\end{defn}
	Let $P=\sum_{i}w_{i}\boldsymbol{x}^{\boldsymbol{\alpha}_{i}}$ be
	a polynomial in $\mathbb{R}\left[\boldsymbol{x}\right]$. Then $\boldsymbol{\alpha}_{i}$
	is called the \emph{coefficient }of the monomial $\boldsymbol{x}^{\boldsymbol{\alpha}_{i}}$.
	If $w_{i}\neq0$, then $w_{i}\boldsymbol{x}^{\boldsymbol{\alpha}_{i}}$
	is called a \emph{term} of $P$. The \emph{maximal degree\index{maximal degree@maximal\emph{ }degree}}
	of $P$, denoted $deg\left(P\right)$, is the maximum $\left|\boldsymbol{\alpha}_{i}\right|$
	such that the coefficient $\alpha_{i}$ is nonzero. For instance,
	the polynomial $P\left(\boldsymbol{x}\right)=5x_{1}^{2}+3x_{1}x_{2}+x_{2}^{2}+x_{1}+x_{2}+3$
	for $\boldsymbol{x}\in\mathbb{R}^{2}$ has six terms and maximal degree
	two. Note that, since $\left|\boldsymbol{\alpha}_{i}\right|$ is defined
	as the sum of monomial degree, so polynomial $P^{\prime}\left(\boldsymbol{x}\right)=5x_{1}^{2}x_{2}^{2}+3x_{1}x_{2}$
	has degree four.
	
	The number of possible monomial terms of a degree $K$ polynomial
	is equivalent to the number of ways of selecting $K$ variables from
	the multisets of $D+1$ variables\footnote{There are $D+1$ variables considering polynomials in homogeneous
		coordinates, i.e. \emph{projective space} $\mathbb{P}^{D}$.}. This is equivalent to the \emph{size $K$ combinations of $D+1$
		elements taken with replacement}. In other words, selecting $K$ variables
	from the variable set $\left(x_{0},x_{1},\ldots,x_{D}\right)$ in
	homogeneous coordinates with repetition, leads to the following fact.
	\begin{fact}
		\emph{If polynomial $P$ in $\mathbb{R}\left[x_{1},x_{2}\ldots,x_{D}\right]$
			has maximal degree $K$, then polynomial $P$ has $\left(\begin{array}{c}
				D+K\\
				D
			\end{array}\right)$ monomial terms at most.\label{fact_G}}
	\end{fact}
	\begin{lem*}
		\emph{For a set points $\mathcal{D}=\left\{ x_{n}\in\mathbb{R}^{D}:n\in\mathcal{N}\right\} $
			in general position, the total number of linear dichotomies in Cover's
			function counting theorem, is the same as the number of cells of the
			dual arrangement $\mathcal{H_{D}}$, plus the number of bounded cells
			of $\mathcal{H_{D}}$. In other words, denote the number of dichotomies
			for $N$ data items in $\mathbb{R}^{D}$ as $\mathrm{Cover}\left(N,D+1\right)$
			($D+1$ denote the dimension of data in homogeneous coordinates) and
			the number of cells and bounded cells of an hyperplane arrangement
			in $\mathbb{R}^{D}$ as $B_{D}\left(\mathcal{H_{D}}\right)$ and $C_{D}\left(\mathcal{H_{D}}\right)$.
			Then 
			\begin{equation}
				\mathrm{Cover}\left(N,D+1\right)=B_{D}\left(\mathcal{H_{D}}\right)+C_{D}\left(\mathcal{H_{D}}\right)
			\end{equation}
		}
	\end{lem*}
	\begin{proof}
		Given a set of\emph{ }points\emph{ $\mathcal{D}=\left\{ x_{n}\in\mathbb{R}^{D}:n\in\mathcal{N}\right\} $}
		in general position.\emph{ }\citealp{cover1965geometrical}'s function
		counting theorem states that the number of linearly separable dichotomies
		given by affine hyperplanes is
		\begin{equation}
			\mathrm{Cover}\left(N,D+1\right)=2\sum_{d=0}^{D}\left(\begin{array}{c}
				N-1\\
				d
			\end{array}\right).
		\end{equation}
		The original Cover's function counting theorem counts the number of
		linearly separable dichotomies given by \emph{linear} hyperplanes.
		However, the dual arrangement\emph{ $\mathcal{H_{D}}$ }consists of
		a set of \emph{affine} hyperplanes\emph{.} Nevertheless, the number
		of dichotomies given by affine hyperplanes in $\mathbb{R}^{D}$ for
		dataset $\mathcal{D}$ is equivalent to the number of dichotomies
		given by linear hyperplanes for dataset $\bar{\mathcal{D}}$ in $\mathbb{R}^{D+1}$
		(where $\bar{\mathcal{D}}$ is the \emph{homogeneous  dataset}, which
		is obtained by embedding $\mathcal{D}$ in homogeneous space. Recall
		that, $\bar{\boldsymbol{x}}=\left(\boldsymbol{x},1\right)$ is the
		data in homogeneous coordinates).
		
		\citet{doi:10.1080/0025570X.1978.11976715} (and see also \citealp{edelsbrunner1986constructing})
		show that shows that for a simple arrangement $\mathcal{H}=\left\{ h_{n}:n\in\mathcal{N}\right\} $
		in $\mathbb{R}^{D}$, the\emph{ }number of cells is $C_{D}\left(\mathcal{H}\right)=\sum_{d=0}^{D}\left(\begin{array}{c}
			N\\
			d
		\end{array}\right)$, and the number of bounded regions is $B_{D}\left(\mathcal{H}\right)=\left(\begin{array}{c}
			N-1\\
			D
		\end{array}\right)$.
		
		Putting these two pieces of information together, obtains
		
		\begin{equation}
			\begin{aligned} & B_{D}\left(\mathcal{H_{D}}\right)+C_{D}\left(\mathcal{H_{D}}\right)\\
				= & \left(\begin{array}{c}
					N-1\\
					D
				\end{array}\right)+\sum_{d=0}^{D}\left(\begin{array}{c}
					N\\
					d
				\end{array}\right)\\
				= & \left(\begin{array}{c}
					N-1\\
					D
				\end{array}\right)+\sum_{d=0}^{D}\left[\left(\begin{array}{c}
					N-1\\
					d
				\end{array}\right)+\left(\begin{array}{c}
					N-1\\
					d-1
				\end{array}\right)\right]\\
				= & \sum_{d=0}^{D}\left(\begin{array}{c}
					N-1\\
					d
				\end{array}\right)+\sum_{d=0}^{D}\left(\begin{array}{c}
					N-1\\
					d
				\end{array}\right)\\
				= & 2\sum_{d=0}^{D}\left(\begin{array}{c}
					N-1\\
					d
				\end{array}\right)\\
				= & \mathrm{Cover}\left(N,D+1\right).
			\end{aligned}
		\end{equation}
	\end{proof}
	\begin{lem*}
		\emph{For a dataset $\mathcal{D}$ in general position, each of Cover's
			dichotomies corresponds to a cell in the dual space, and dichotomies
			corresponding to bounded cells have no }complement cell\emph{ (cells
			with reverse sign vector). Dichotomies corresponding to the unbounded
			cells in the dual arrangements $\phi\left(\mathcal{D}\right)$ have
			a complement cell.}
	\end{lem*}
	\begin{proof}
		The first statement is true because of the order preservation property
		-- data item $\boldsymbol{x}$ lies above (below) hyperplane $h$
		if and only if point $\phi^{-1}\left(h\right)$ lies above (below)
		hyperplane $\phi\left(\boldsymbol{x}\right)$. For a dataset $\mathcal{D}$
		and hyperplane $h$, assume $h$ has a normal vector $\boldsymbol{w}$
		(in homogeneous coordinates) and there is no data item lying on $h$.
		Then, hyperplane $h$ will partition the set $\mathcal{D}$ into two
		subsets $\mathcal{D}_{h}^{+}=\left\{ \boldsymbol{x}_{n}:\boldsymbol{w}^{T}\boldsymbol{x}>0\right\} $
		and $\mathcal{D}_{h}^{-}=\left\{ \boldsymbol{x}_{n}:\boldsymbol{w}^{T}\boldsymbol{x}<0\right\} $,
		and according to the Thm. \ref{Theorem 6. The Incidence relations of dual transformation},
		$\mathcal{D}$ has a unique associated dual arrangement $\phi\left(\mathcal{D}\right)$.
		Thus, the sign vector of the point $\phi^{-1}\left(h\right)$ with
		respect to arrangement $\phi\left(\mathcal{D}\right)$ partitions
		the arrangement into two subsets $\phi_{h}\left(\mathcal{D}\right)^{+}=\left\{ \phi\left(\boldsymbol{x}_{n}\right):\boldsymbol{\nu}_{\phi\left(\boldsymbol{x}_{n}\right)}^{T}\phi^{-1}\left(h\right)>0\right\} $
		and $\phi_{h}\left(\mathcal{D}\right)^{-}=\left\{ \boldsymbol{\nu}_{\phi\left(\boldsymbol{x}_{n}\right)}^{T}\phi^{-1}\left(h\right)<0\right\} $,
		where $\boldsymbol{\nu}_{\phi\left(\boldsymbol{x}_{n}\right)}^{T}$
		is the normal vector to the dual hyperplane $\phi\left(\boldsymbol{x}_{n}\right)$,
		in other words, point $\phi^{-1}\left(h\right)$ lies in a cell of
		arrangement $\phi\left(\mathcal{D}\right)$.
		
		Next, it is necessary to prove that bounded cells have no complement
		cell. The reverse assignment of the bounded cells of the dual arrangements
		$\phi\left(\mathcal{D}\right)$ cannot appear in the primal space
		since the transformation $\phi$ can only have normal vector $\boldsymbol{\nu}$
		pointing in one direction, in other words, transformation $\phi$:
		$x_{D}=p_{1}x_{1}+p_{2}x_{2}+...+p_{D-1}x_{D-1}-p_{D}$ implies the
		$D$th component of normal vector $\boldsymbol{\nu}$ is $-1$. For
		unbounded cells, in dual space, every unbounded cell $f$ associates
		with another cell $g$, such that $g$ has an opposite sign vector
		to $f$. This is because every hyperplane $\phi\left(\boldsymbol{x}_{n}\right)$
		is cut by another \emph{$N-1$} hyperplanes into \emph{$N+1$} pieces
		(since in a simple arrangement no two hyperplanes are parallel), and
		each of the hyperplanes contains two rays, call them $\boldsymbol{r}_{1},$$\boldsymbol{r}_{2}$.
		These two rays point in opposite directions, which means that the
		cell incident with $\boldsymbol{r}_{1}$ has an opposite sign vector
		to $\boldsymbol{r}_{2}$ with respect to all other \emph{$N-1$} hyperplanes.
		Therefore, it is only necessary to take the cell $f$ incident with
		$\boldsymbol{r}_{1}$, and in the positive direction with respect
		to $\phi\left(\boldsymbol{x}_{n}\right)$, take $g$ to be the cell
		incident with $\boldsymbol{r}_{2}$, and in the negative direction
		with respect to $\phi\left(\boldsymbol{x}_{n}\right)$. In this way,
		two unbounded cells $f$ and $g$ are obtained with opposite sign
		vectors. This means that, for point $\phi^{-1}\left(h\right)$ in
		these unbounded cells, this hyperplane $h$ partitions the dataset
		to $\mathcal{D}_{h}^{+}$ and $\mathcal{D}_{h}^{-}$, and it is possible
		to move the position of hyperplane $h$ in the primal space. So, there
		exists a new hyperplane $h^{\prime}$ obtained by moving $h$, and
		it partitions the dataseto $\mathcal{D}_{h^{\prime}}^{+}=\mathcal{D}_{h}^{-}$
		and $\mathcal{D}_{h^{\prime}}^{-}=\mathcal{D}_{h}^{+}$. In other
		words, $h^{\prime}$ has opposite assignment compared to hyperplane
		$h$. This corresponds, in the dual space, to moving a point $\phi^{-1}\left(h\right)$
		inside the cell $f$, to cell $g$. For instance, in the simplest
		case, a hyperplane can be moved from left-most to the right-most to
		obtain an opposite assignment without changing the direction of the
		normal vector.
	\end{proof}
	Since each of Cover's dichotomies corresponds to a cell in the dual
	space, and dichotomies corresponding to bounded cells have no complement
	cell (cells with reverse sign vector), lemma \ref{Lemma: dichotomies  and cells}
	demonstrates that all possible\emph{ Cover's dichotomies }of a given
	dataset $\mathcal{D}$ can be obtained by enumerating the cells of
	an arrangement and the complemented cells of the bounded cells. The
	enumeration of the complements of the bounded cells requires an additional
	process, as the bounded cells within the arrangement do not have complementary
	cells. This result directly leads to the following theorem.
	\begin{lem*}
		\emph{For a dataset $\mathcal{D}$ in general position, a hyperplane
			with $k$ data items lying on it, $0\leq k\leq D$ correspond to a
			$\left(D-k\right)$-face in the dual arrangement $\mathcal{H_{D}}$.
			Hyperplanes with $D$ points lying on it, correspond to vertices in
			the dual arrangement.}
	\end{lem*}
	\begin{proof}
		According to the incidence preservation property, $k$ data items
		lying on a hyperplane will intersect with $k$ hyperplanes, and the
		intersection of $k$ hyperplanes will create a ($D-k$)-dimensional
		space, which is a $\left(D-k\right)$-face, and the $0$-faces are
		the \emph{vertices} of the arrangement.
	\end{proof}
	\begin{lem*}
		\emph{Given a hyperplane arrangement $\mathcal{H}=\left\{ h_{n}:n\in\mathcal{N}\right\} $,
			for an arbitrary maximal face (cell) $f$, the sign vector of $f$
			is $\text{sign}_{\mathcal{H}}\left(f\right)$. For an arbitrary $\left(D-d\right)$-dimension
			face $g$, $0<d\leq D$, the number of different signs of $\text{sign}_{\mathcal{H}}\left(g\right)$
			with respect to $\text{sign}_{\mathcal{H}}\left(f\right)$ is larger
			than or equal to $d$, where equality holds only when $g$ is conformal
			to $f$ ($g$ is a subface of $f$).}
	\end{lem*}
	\begin{proof}
		Denote the number of different signs of\emph{ }$\text{sign}_{\mathcal{H}}\left(g\right)$
		with respect to $\text{sign}_{\mathcal{H}}\left(f\right)$ by\emph{
			$E_{\text{0-1}}\left(g\right)$.} In a simple arrangement, the sign
		vector $\text{sign}_{\mathcal{H}}\left(f\right)$ of a cell $f$ has
		no zero signs, and a $\left(D-d\right)$-dimension face has $d$ zero
		signs. Thus the number of different signs of\emph{ }$\text{sign}_{\mathcal{H}}\left(g\right)$
		with respect to\emph{ }$\text{sign}_{\mathcal{H}}\left(f\right)$
		must be larger than or equal to \emph{$d$}, i.e., $E_{\text{0-1}}\left(f\right)\geq d$.
		If $\mathrm{sep}\left(f,g\right)=\emptyset$, then $E_{\text{0-1}}\left(g\right)=d$
		according to the definition of $\mathrm{sep}\left(f,g\right)=\emptyset$.
		In this case, $f$, $g$ are conformal\emph{.} By contrast, if $f$,
		$g$ are not conformal, i.e. $\mathrm{sep}\left(f,g\right)\neq\emptyset$,
		and assuming $\left|\mathrm{sep}\left(f,g\right)\right|=C$, then
		according to the definition of the objective function and conformal
		faces, $E_{\textrm{0-1}}\left(\text{sign}_{\mathcal{H}}\left(f\right)\right)=d+C$.
		Hence, $E_{\textrm{0-1}}\left(\text{sign}_{\mathcal{H}}\left(g\right)\right)\geq d$,
		and equality holds only when $g$ is conformal to $f$.
	\end{proof}
	\begin{thm*}
		\emph{Consider a  dataset $\mathcal{D}$ of $N$ data points of dimension
			$D$ in general position, along with their associated labels, denote
			$\mathcal{S}_{\text{kcombs}}$ as the set of all $D$-combinations
			with respect to  dataset $\mathcal{D}$. Then we have following inequality
			\begin{equation}
				\underset{s\in\mathcal{S}_{\text{kcombs}}}{\text{argmin}}\min\left(E_{\text{0-1}}\left(\boldsymbol{w}_{s}\right),E_{\text{0-1}}\left(-\boldsymbol{w}_{s}\right)\right)\subseteq\underset{\boldsymbol{w}\in\mathbb{R}^{D+1}}{\text{argmin}}E_{\text{0-1}}\left(\boldsymbol{w}\right)\label{eq: 0-1 loss classification theorem-1}
			\end{equation}
			where $\boldsymbol{w}_{s}$ represents the normal vector of the hyperplane
			that pass through the $D$-combination of data $s$, and $-\boldsymbol{w}_{s}$
			is the negation of $\boldsymbol{w}_{s}$. The inner $\min$ on the
			left-hand side ensures that for each $s\in\mathcal{S}_{\text{kcombs}}$,
			we take the smaller of $E_{\text{0-1}}\left(\boldsymbol{w}_{s}\right)$
			and $E_{\text{0-1}}\left(-\boldsymbol{w}_{s}\right)$, the outer $\text{argmin}$
			finds }one\emph{ of the value of that minimizes this quantity over
			all $s\in\mathcal{S}_{\text{kcombs}}$. In other words, (\ref{eq: 0-1 loss classification theorem-1})
			means that all globally optimal solutions to problem (\ref{eq:0-1-loss-linear-classification-problem}),
			are equivalent (in terms of 0-1 loss) to the optimal solutions contained
			in the set of solutions of all positive and negatively-oriented linear
			classification decision hyperplanes (vertices in the dual space) which
			go through $D$ out of $N$ data points in the  dataset $\mathcal{D}$.
			\label{0-1 loss classification theorem}}
	\end{thm*}
	\begin{proof}
		First, transform a  dataset$\text{\ensuremath{\mathcal{D}}}$ to its
		dual arrangement. According to Lemma \ref{Lemma: dichotomies  and cells}
		and Lemma \ref{vertex-hyperplane-connection}, each dichotomy has
		a corresponding dual cell and if the sign vectors for all possible
		cells in the dual arrangement and their reverse signs are evaluated,
		the optimal solution for the 0-1 loss classification problem can be
		obtained. Assume the optimal cell is $f$, it is required to prove
		that, one of the adjacent vertices for this cell is also the optimal
		vertex. Then, finding an optimal vertex is equivalent to finding an
		optimal cell since the optimal cell is one of the adjacent cells of
		this vertex. According to Lemma \ref{optimal conformal faces}, any
		vertices that are non-conformal have corresponding 0-1 loss with respect
		to $\text{sign}_{\mathcal{H}}\left(f\right)$ which is strictly greater
		than $D$. Since $f$ is optimal, any sign vectors with larger sign
		difference (with respect to $\text{sign}_{\mathcal{H}}\left(f\right)$)
		will have larger 0-1 loss value (with respect to true label $\boldsymbol{t}$).
		Therefore, vertices that are conformal to $f$ will have smaller 0-1
		loss value, thus one can evaluate all vertices (and the reverse sign
		vector for these vertices) and choose the best one, which, according
		to Lemma \ref{vertex-hyperplane-connection}, is equivalent to evaluating
		all possible positive and negatively-oriented linear classification
		decision hyperplanes and choosing one linear decision boundary with
		the smallest 0-1 loss value.
	\end{proof}
	\begin{theorem}
		Symmetry fusion theorem\emph{. Consider a  dataset $\mathcal{D}$ of
			$N$ data points of dimension $D$ in general position, along with
			their associated labels. Let $h$ be a hyperplane which goes through
			$D$ out of $N$ data points in the  dataset $\mathcal{D}$, separating
			the  dataset into two disjoint sets $\mathcal{D}^{+}$ and $\mathcal{D}^{-}$.
			If the 0-1 loss for the positive orientation of this hyperplane is
			$l$, then the 0-1 loss for the negative orientation of this hyperplane
			is $N-l-D$. }
	\end{theorem}
	\begin{proof}
		Assume there are $m^{+}$ and $m^{-}$ data points misclassified in
		\emph{$\mathcal{D}^{+}$ }and \emph{$\mathcal{D}^{-}$}, then the
		0-1 loss for $h$ equals $l=m^{+}+m^{-}$\emph{.} Denote the hyperplane
		$h$ with negative orientation as $h^{-}$. In the partition introduced
		by $h^{-}$, all correctly classified data by $h$ will be misclassified
		in $h^{-}$. Thus the 0-1 loss of $h^{-}$ is $\left|\mathcal{D}^{+}\right|-m^{+}+\left|\mathcal{D}^{-}\right|-m^{-}$.
		Since $\left|\mathcal{D}^{+}\right|+\left|\mathcal{D}^{-}\right|=N-D$,
		we obtain the 0-1 loss for $h^{-}$ which is $N-D-l$.
	\end{proof}
	
	\section{Addtional experiments}\label{sec:Additional experiments}
	
	\begin{table*}
		\centering
			\begin{tabular}{l@{\hskip 4pt}c@{\hskip 4pt}cp{2cm} p{2cm} p{2cm} p{2cm}}
				\toprule
				 datasets & $N$ & $D$ & ICE (\%) & SVM (\%) & LR (\%) & LDA (\%) \tabularnewline
				\midrule
				
				HA & 283 & 3 & *\textbf{77.35}/\textbf{74.39}
				
				(0.58)/(2.66) & 72.21/71.23
				
				(0.58)/(2.00) & 72.39/72.28
				
				(0.92)/(0.15) & 73.10/74.74
				
				(0.85)/(2.66)\tabularnewline
				\hline 
				CA & 72 & 5 &* \textbf{81.75}/\textbf{61.33}
				
				(2.66)/(5.58) & 71.93/60.00
				
				(7.85)/(0816) & 76.49/58.67
				
				(4.40)/(7.30) & 76.14/58.67
				
				(6.75)/(7.30)\tabularnewline
				\hline 
				CR & 89 & 6 & *\textbf{95.49}/83.33
				
				(1.18)/(7.86) & 92.11/\textbf{85.56}
				
				(1.89)/(10.09) & 90.99/82.22
				
				(2.36)/(9.13) & 90.99/82.22
				
				(2.36)/(12.67)\tabularnewline
				\hline 
				VP & 704 & 2 & *\textbf{96.93}/\textbf{97.59}
				
				(0.44)/(0.15) & 96.77/97.02
				
				(0.49)/(2.32) & 96.02/96.03
				
				(0.00)/(0.29) & 96.48/96.88
				
				(0.63)/(2.28)\tabularnewline
				\hline 
				BT & 502 & 4 & *\textbf{79.50}/\textbf{74.06}
				
				(0.82)/(2.36) & 74.96/72.67
				
				(0.82)/(2.85) & 76.06/73.27
				
				(0.79)/(3.50) & 75.81/73.07
				
				(1.09)/(3.53)\tabularnewline
				\hline 
				SP & 975 & 3 & *\textbf{94.49}/\textbf{94.15}
				
				(0.27)/(1.00) & 94.13/93.74
				
				(0.33)/(1.33) & 94.13/93.74
				
				(0.33)/(0.13) & 94.13/93.74
				
				(0.33)/(1.33)\tabularnewline
				\hline 
				Ai4i & 10000 & 6 & \textbf{97.45}/\textbf{97.40}
				
				(0.10)/(0.36) & 96.62/96.57
				
				(0.33)/(0.53) & 96.99/96.90
				
				(0.10)/(0.44) & 97.00/96.75
				
				(0.13)/(0.33)\tabularnewline
				\hline 
				AIDS & 2139 & 23 & \textbf{87.75}/\textbf{87.61}
				
				(1.09)/(1.12) & 86.84/86.49
				
				(0.32)/(1.24) & 86.56/86.58
				
				(0.19)/(1.23) & 85.71/84.90
				
				(0.25)/(1.30)\tabularnewline
				\hline 
				AL & 243 & 14 & \textbf{98.45}/\textbf{98.36}
				
				(0.92)/(2.37) & 95.77/95.10
				
				(0.89)/(3.05) & 96.18/95.51
				
				(1.06)/(4.16) & 94.53/88.57
				
				(0.53)/(4.21)\tabularnewline
				\hline 
				AV & 2043 & 7 & \textbf{88.94}/\textbf{88.31}
				
				(0.25)/(2.21) & 87.14/87.33
				
				(0.41)/(1.64) & 86.94/87.04
				
				(0.36)/(1.51) & 86.27/86.70
				
				(0.42)/(1.39)\tabularnewline
				\hline 
				RC & 3810 & 7 & \textbf{93.86}/92.55
				
				(0.29)/(1.05) & 92.83/\textbf{92.78}
				
				(0.18)/(0.78) & 92.86/92.81
				
				(0.25)/(0.67) & 93.14/92.65
				
				(0.17)/(0.59)\tabularnewline
				\hline 
				DB & 1146 & 19 & \textbf{79.48}/\textbf{79.74}
				
				(1.76)/(0.70) & 69.63/67.65
				
				(0.01)/(0.03) & 70.57/69.39
				
				(0.01)/(0.03) & 73.93/70.61
				
				(0.01)/(0.02)\tabularnewline
				\hline 
				SO & 1941 & 27 & \textbf{77.70/76.04}
				
				(0.45)/(0.84) & 73.03/73.62
				
				(0.01)/(0.01) & 72.78/72.96
				
				(0.00)/(0.02) & 73.81/74.81
				
				(0.01)/(0.03)\tabularnewline
				\hline 
				SS & 51433 & 3 & \textbf{86.58/86.68}
				
				(0.04)/(0.19) & 82.78/82.71
				
				(0.00)/(0.00) & 79.69/79.65
				
				(0.00)/(0.00) & 80.31/80.32
				
				(0.00)/(0.00)\tabularnewline
				
				\bottomrule
			\end{tabular}
			\caption{Comparison of the accuracy of our novel incremental
				cell enumeration (ICE) algorithm, against approximate methods: SVM,
				logistic regression (LR), and linear discriminant analysis (LDA) on real-world
				 datasets. Results are reported as mean accuracy loss over training and test sets in the format: Training Error / Test Error (Standard Deviation: Train / Test). Exact solutions are marked with *, otherwise approximate, obtained using stochastic coreset selection for tractability purposes (\ref{sec:Coreset-selection-method}). Best performing algorithm is marked bold. }
			\label{tab:Additional Oos}
		\end{table*}
		
		\begin{figure}
			\begin{centering}
				\includegraphics[width=3.5in]{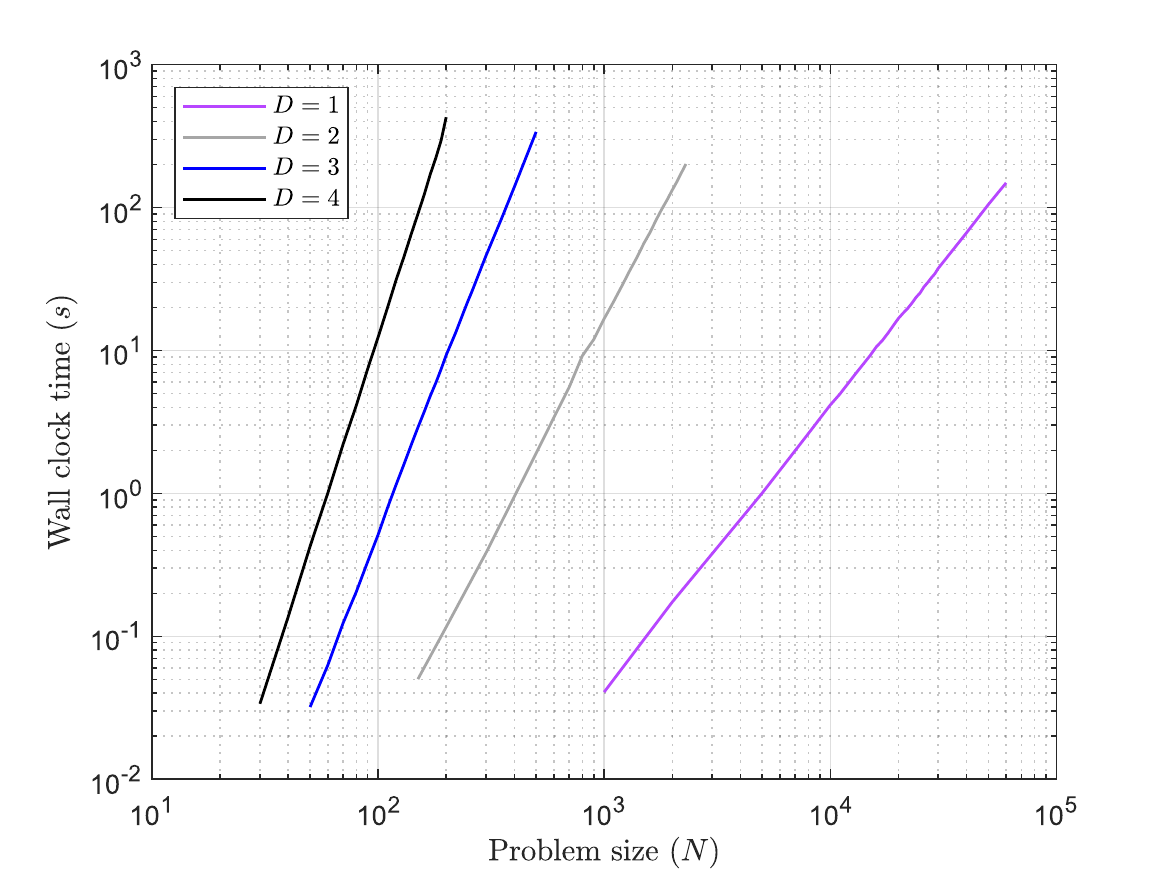}
				\par\end{centering}
			\caption{Log-log wall-clock run time (seconds) for the ICE algorithm in $1D$
				to $4D$ synthetic  datasets, against  dataset size $N$, where the
				approximate upper bound is disabled (by setting it to $N$). The run-time
				curves from left to right (corresponding to $D=1,2,3,4$ respectively),
				have slopes 2.0, 3.1, 4.1, and 4.9, a very good match to the predicted
				worst-case run-time complexity of $O\left(N^{2}\right)$, $O\left(N^{3}\right)$,
				$O\left(N^{4}\right)$, and $O\left(N^{5}\right)$ respectively. \label{fig:run-time-polynomial}}
		\end{figure}
		
		\begin{figure}
			\begin{centering}
				\includegraphics[width=3.5in]{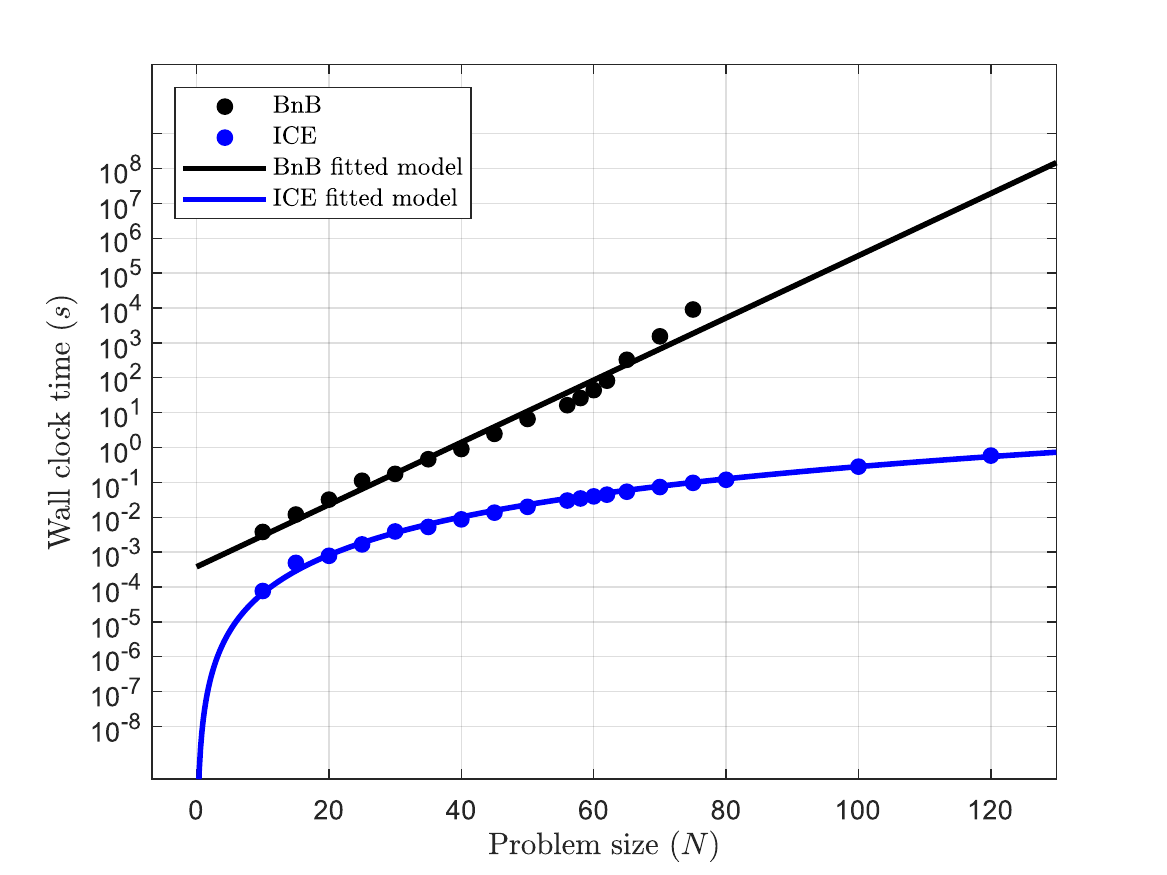}
				\par\end{centering}
			\caption{Log-linear wall-clock run time (seconds) plot comparing the ICE algorithm
				against the branch-and-bound (BnB) algorithm of \citet{SIAM-v28-nguyen13a}
				(Matlab implementation provided by the authors) on three dimensional
				synthetic data. On this log-linear scale exponential run time appears
				as a linear function of problem size $N$, whereas, polynomial run
				time is a logarithmic function of $N$. Fitting appropriate models
				(lines) to the computational experiment data (dots) provides clear
				evidence of this prediction.\label{fig:ice-vs-bnb}}
		\end{figure}
		
		\subsection{Run-time complexity analysis\label{sec:Run-time-complexity-analysis}}
		
		We test the wall clock time of our novel ICE algorithm on four different
		synthetic datasets with dimension ranging from $1D$ to $4D$. The
		1$D$-dimensional dataset has data size ranging from $N=1000$ to
		60000, the 2$D$-dimensional ranges from 150 to 2400, 3$D$-dimensional
		from 50 to 500, and 4$D$-dimensional data ranging from 30 to 200.
		The worst-case predictions are well-matched empirically (see Figure
		\ref{fig:run-time-polynomial}).

		All decision boundaries computed by exact algorithms entail the same,
		globally optimal 0-1 loss. Therefore, the only meaningful comparison
		between ICE and any other exact algorithms is in terms of time complexity.
		Here, we compare the wall-clock run time of our ICE algorithm with
		the exact \emph{branch-and-bound} (BnB) algorithm of \citet{SIAM-v28-nguyen13a}.
		As a branch-and-bound algorithm, in the worst case it must test all
		possible assignments of data points to labels which requires an exponential
		number of computations, by comparison to ICE's worst case polynomial
		time complexity arising from the enumeration of dichotomies instead.
		Empirical computations confirm this reasoning (see Figure \ref{fig:ice-vs-bnb}),
		predicting for instance that for the $N=150$ data size with $D=3$,
		ICE would take \textbf{1.2 seconds} worst-case whereas BnB would take
		approximately $10^{10}$ seconds (nearly \textbf{317 years}), demonstrating
		the clear superiority of our approach.
	
\begin{figure}
		\begin{centering}
			\includegraphics[width=3.5in]{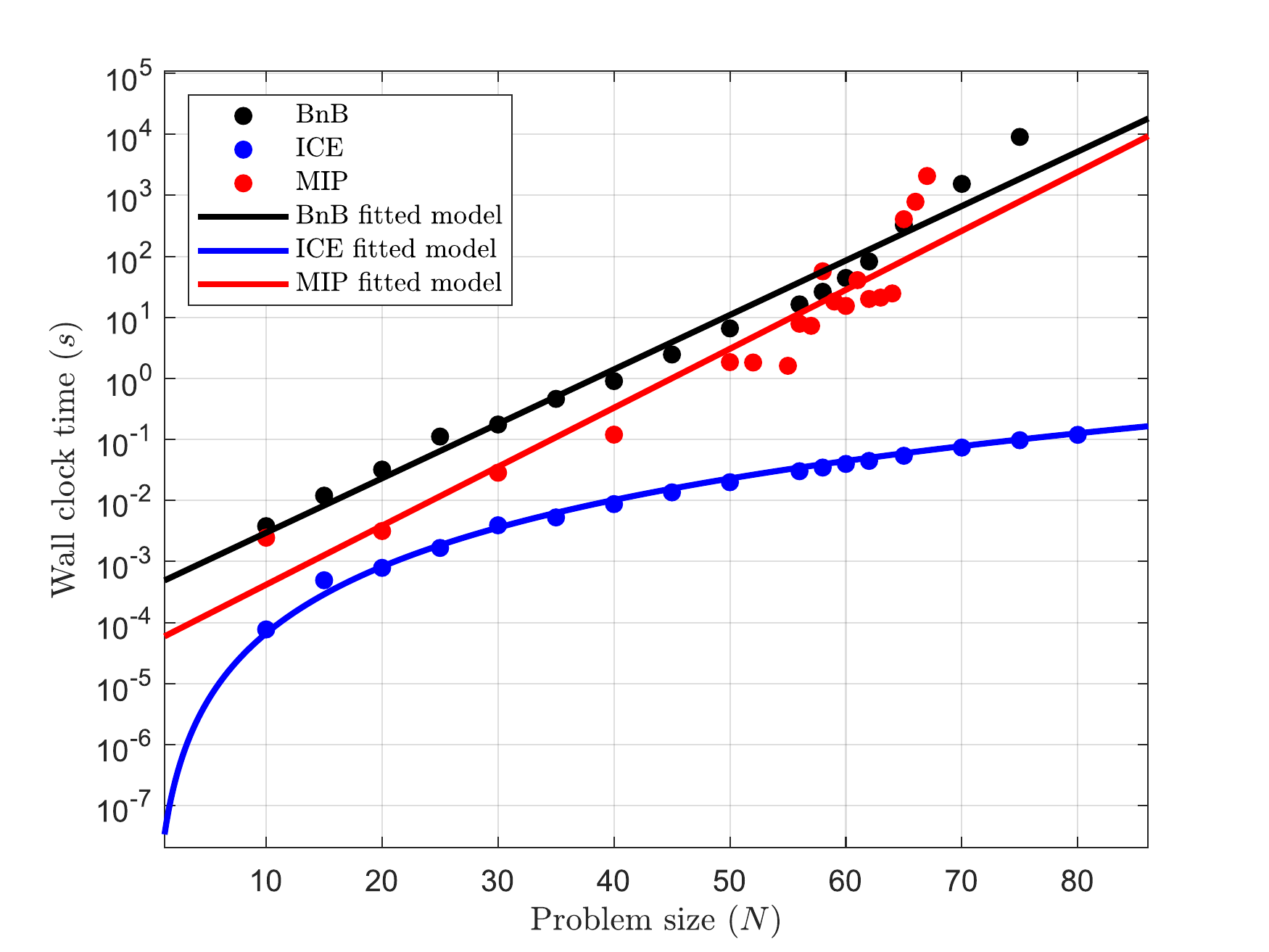}
			\par\end{centering}
		\caption{Log-linear wall-clock run time (seconds) plot comparing the ICE algorithm
			against the branch-and-bound (BnB) algorithm of \citet{SIAM-v28-nguyen13a}
			(MATLAB implementation provided by the authors) and the mixed-integer
			programming (MIP) solver (implemented in MATLAB using GLPK solver)
			on three dimensional synthetic data. On this log-linear scale exponential
			run time appears as a linear function of problem size $N$, whereas,
			polynomial run time is a logarithmic function of $N$. Fitting appropriate
			models (lines) to the computational experiment data (dots) provides
			clear evidence of this prediction. The smaller sampling size of the
			MIP solver compared with BnB and ICE is due to the solver not terminating
			within the three-hour time limit, highlighting its much less predictable
			performance. \label{fig: ice-vs-bnb-mip}}
	\end{figure}

	Additionally, we compare the performance of the ICE and BnB algorithms
	with that of a mixed-integer programming (MIP) solver for the 0-1
	LCP, the results is shown
	in Figure \ref{fig: ice-vs-bnb-mip}. The results show that while
	the MIP solver is more efficient than BnB on small  datasets, its performance
	is less predictable compared with ICE and BnB. This is highlighted
	by the fact that the number of sampling points explored by the MIP
	solver is smaller than those of BnB and ICE, as the MIP solver had
	not yet terminated to obtain the exact solution within our three-hour
	time limit.

		\subsection{Out-of-Sample Generalization Test}

Due to the inherent combinatorial complexity of the 0-1 loss classification
problem, the ICE algorithm becomes computationally intractable for
high-dimensional  datasets. In Table \ref{tab:Additional Oos} , except
for  datasets that are tractable for ICE algorithm (those  datasets
evaluated in Table \ref{tab:empirical-error-comparisons}), we train
all other  datasets using the coreset selection method (ICE-coreset)
introduced by \citep{he2025deepicegloballyoptimalalgorithm}. This
method acts as a randomized wrapper for exact algorithms (pseudocode
is provided in Appendix \ref{sec:Coreset-selection-method}). In brief,
the coreset selection method reduces the  dataset by eliminating subsets
associated with solutions exhibiting higher 0-1 loss. This process
is repeated until the reduced  dataset (the coreset) becomes tractable
for the ICE algorithm. As predicted, the solutions obtained by the
ICE-coreset algorithm not only perform well on training data but also
demonstrate higher test accuracy, refuting the misconception that
exact algorithms necessarily overfit the training set.
		
		\subsection{Hypersurface classification}
		The out-of-sample generalization performance on real-world  datasets for the quadratic classifier is reported in Table \ref{tab:quadratic compare}.
			\begin{table*}
			\centering
				\begin{tabular}{l@{\hskip 4pt}c@{\hskip 4pt}cp{2cm} p{2cm}}
					\toprule
					datasets & $N$ & $D$ & ICE (\%) & SVM (\%)  \tabularnewline
					\midrule

					HA & 283 & 3 & \textbf{77.35}/\textbf{73.68}
					
					(0.20/0.00)& 73.01/66.67
					
					(0.00/0.00)\tabularnewline
	 
					CA & 72 & 5 & \textbf{78.95}/\textbf{86.66}
					
					(0.01/0.00) & 70.18/46.67
					
					(0.02/0.01)\tabularnewline
	
					CR & 89 & 6 & \textbf{92.96}/\textbf{94.44}
					
					0.20/0.13 & 91.55/\textbf{94.44}
					
					(0.00/0.00)\tabularnewline
		
					VP & 704 & 2 & \textbf{97.12}/\textbf{96.45}
					
					(0.01/0.00) & 96.63/95.74
					
					(0.00/0.01)\tabularnewline
		
					BT & 502 & 4 & \textbf{78.30}/\textbf{80.00}
					
					0/0.44 & 73.56/75.25
					
					(0.01/0.02)\tabularnewline
		
					SP & 975 & 3 & \textbf{95.13}/\textbf{91.79}
					
					(0.00/0.00) & 94.74/91.28
					
					(0.00/0.00)\tabularnewline
		
				\end{tabular}

			\caption{Empirical comparison of the training accuracy of our ICE algorithm
				against an approximate SVM with a degree-2 polynomial kernel on real-world
				datasets. Results are reported as mean accuracy over training and
				test sets in the format: Training Accuracy / Test Accuracy (Standard
				Deviation: Train / Test). Best performing algorithm is marked bold.
				\label{tab:quadratic compare}}
		\end{table*}
		
		\section{Coreset selection method\label{sec:Coreset-selection-method}}
		
		Algorithm \ref{alg:ice-coreset} shows the structure of the coreset selection method.
		
		\begin{algorithm}[tb]
			\caption{ICE with coreset filtering}
			\label{alg:ice-coreset}
			\textbf{Input}: $M$: block size; $R$: number of shuffles in each filtering round; $L$: max-heap size; $B_{\max}$: maximum input size for ICE algorithm; $c \in (0,1]$: heap shrinking factor \\
			\textbf{Output}: Max-heap $\mathcal{H}_L$ containing top $L$ configurations and associated data blocks
			\begin{algorithmic}[1]
				\STATE $\mathcal{C} \gets ds$ \textit{// initialize coreset with dataset}
				\WHILE{$|\mathcal{C}| \leq B_{\max}$}
				\STATE Divide $\mathcal{C}$ into $\left\lceil \frac{|\mathcal{C}|}{M} \right\rceil$ blocks: $\mathcal{C}_B = \{C_1, C_2, \dots, C_{\left\lceil \frac{|\mathcal{C}|}{M} \right\rceil}\}$
				\STATE Initialize max-heap $\mathcal{H}_L$ of size $L$
				\FOR{$r = 1$ \TO $R$}
				\FORALL{$C \in \mathcal{C}_B$}
				\STATE $\mathit{cnfg} \gets \mathit{ICE}(\mathcal{D}_{\boldsymbol{l}}, K)$
				\STATE $\mathcal{H}_L.\text{push}(\mathit{cnfg}, C)$
				\ENDFOR
				\STATE $\mathcal{C} \gets \mathit{unique}(\mathcal{H}_L)$ \textit{// merge blocks and remove duplicates}
				\STATE $L \gets L \times c$ \textit{// shrink heap size}
				\ENDFOR
				\ENDWHILE
				\STATE $\mathit{cnfg} \gets \mathit{ICE}(\mathcal{D}_{\boldsymbol{l}}, K)$ \textit{// final refinement}
				\STATE $\mathcal{H}_L.\text{push}(\mathit{cnfg}, \mathcal{C})$
				\STATE \textbf{return} $\mathcal{H}_L$
			\end{algorithmic}
		\end{algorithm}


\end{document}